%% file: ms.tex
\documentclass{article}
\pdfoutput=1

\usepackage[final]{neurips_2022}
\usepackage[utf8]{inputenc}

\title{The Role of Local Alignment and Uniformity in Image-Text Contrastive Learning on Medical Images}
\author{
    Philip M\"uller \\
    Technical University of Munich \\
    \texttt{philip.j.mueller@tum.de} \\ 
    \AND
    Georgios Kaissis\\
    Technical University of Munich \\
    Helmholtz Zentrum Munich \\
    \And
    Daniel Rueckert \\
    Technical University of Munich \\
    Imperial College London \\
}

\usepackage{multirow}
\usepackage{hhline}
\usepackage{arydshln}
\usepackage{booktabs}
\usepackage{dsfont}
\usepackage{placeins}
\usepackage[inline]{enumitem}
\usepackage{pifont}
\usepackage{xcolor, colortbl}
\usepackage[normalem]{ulem}
\newcommand{\cmark}{\ding{51}}
\newcommand{\xmark}{\ding{55}}
\definecolor{Gray}{gray}{0.9}
\definecolor{LightCyan}{rgb}{0.88,1,1}
\usepackage{tikz}
\usepackage{comment}
\usepackage[caption=false]{subfig}
\usepackage{amsmath,amssymb} 
\usepackage{bm}
\usepackage{makecell}
\usepackage[capitalize]{cleveref}
\crefname{section}{Sec.}{Secs.}
\Crefname{section}{Section}{Sections}
\Crefname{table}{Table}{Tables}
\crefname{table}{Tab.}{Tabs.}

\renewcommand{\S}{\mathcal{I}}
\newcommand{\R}{\mathcal{R}}

\newcommand{\xs}{\bm{x}^{\S}_i}
\newcommand{\xr}{\bm{x}^{\R}_i}
\newcommand{\ys}[1][k]{\bm{y}^{\S}_{i,#1}}
\newcommand{\yr}[1][m]{\bm{y}^{\R}_{i,#1}}
\newcommand{\ygs}{\bar{\bm{y}}^{\S}_{i}}
\newcommand{\ygr}{\bar{\bm{y}}^{\R}_{i}}
\newcommand{\zs}[1][k]{\bm{z}^{\S}_{i,#1}}
\newcommand{\zr}[1][m]{\bm{z}^{\R}_{i,#1}}
\newcommand{\zsj}[1][k]{\bm{z}^{\S}_{j,#1}}
\newcommand{\zrj}[1][m]{\bm{z}^{\R}_{j,#1}}
\newcommand{\zgs}[1][i]{\bar{\bm{z}}^{\S}_{#1}}
\newcommand{\zgr}[1][i]{\bar{\bm{z}}^{\R}_{#1}}
\newcommand{\zsr}[1][m]{\bm{z}^{\S \rightarrow \R}_{i,#1}}
\newcommand{\zrs}[1][k]{\bm{z}^{\R \rightarrow \S}_{i,#1}}

\newcommand{\weights}[1][k]{w^{\S}_{i, #1}}

\newcommand{\weightr}[1][m]{w^{\R}_{i, #1}}
\newcommand{\alphars}[1][k]{\alpha^{\R \rightarrow \S}_{i, #1, m}}
\newcommand{\alpharsl}{\alpha^{\R \rightarrow \S}_{i, l, m}}
\newcommand{\alphasr}[1][m]{\alpha^{\S \rightarrow \R}_{i, #1, k}}
\newcommand{\ps}{p^{\S}_{k, l}}
\newcommand{\pslk}{p^{\S}_{l, k}}

\begin{document}

\maketitle
\begin{abstract}
Image-text contrastive learning has proven effective for pretraining medical image models. When targeting localized downstream tasks like semantic segmentation or object detection, additional local contrastive losses that align image regions with sentences have shown promising results. We study how local contrastive losses are related to global (per-sample) contrastive losses and which effects they have on localized medical downstream tasks. Based on a theoretical comparison, we propose to remove some components of local losses and replace others by a novel distribution prior which enforces uniformity of representations within each sample. We empirically study this approach on chest X-ray tasks and find it to be very effective, outperforming methods without local losses on 12 of 18 tasks.%, indicating the importance of such local uniformity.
\end{abstract}

\section{Introduction}
Image-text contrastive learning~\citep{CLIP,ALIGN} has been well established as a pretraining method for image models recently. 
By utilizing companion text like radiological reports or captions, it can improve the downstream performance of image models on tasks like image classification. Such image-text methods have also proven effective on medical images like chest X-rays~\citep{ConVIRT,LoVT_MICCAI}. 
Typically, such methods use (global) contrastive losses to align per-sample representation of images and the related text.
However, when targeting localized downstream tasks like semantic segmentation or object detection, it has proven beneficial to also introduce local contrastive losses to align image regions (e.g.\ patches) with sentences~\citep{LoVT,local_MI}.

In this work we study how local contrastive losses are related to global contrastive losses and which effects they have on localized downstream tasks. Following \citet{uniformity}, we decompose the global and local losses into alignment components (pulling representations close to each other) and distribution priors (pushing representations away from each other). We found that the alignment components of global and local losses are related and assume that they have similar effects, while the distribution priors of global and local losses have complementary effects. We empirically study these findings on localized downstream tasks on chest X-rays and therefore propose a pre-training method consisting of a global contrastive loss and local uniformity regularizers (i.e.\ distribution priors) but without local alignment components. Our results show that this method typically performs well on tasks where local contrastive methods outperform global contrastive methods, thus proving our assumptions and indicating the relevance of local uniformity for localized downstream tasks.
For simplicity, we focus our study on the LoVT method~\citep{LoVT}, but argue that our findings are also relevant for related approaches (see \cref{sec:other_methods}).

%Our contributions are as follows:
%\begin{itemize}
    %\item We study the mathematical relation of the alignment components of global and local contrastive losses
%    \item We decompose the global and local contrastive losses of LoVT\cite{LoVT} into alignment and distribution prior components.
%    \item We reformulate the global and local alignment components in a generalized form and identify their main differences. We found that, while they are not identical, they are highly related such that optimizing one may have similar effects as optimizing both.
    %\item We study the difference between the global and local distribution priors and argue that they have different effects and are complementary. While the global distribution prior imposes a distribution over the dataset, the local prior imposes a distribution within each sample.
    %\item We propose a local distribution prior acting on each modality independently, replacing the distribution prior from the local contrastive losses.
%    \item We empirically study replacing the local contrastive losses of LoVT by the proposed priors while removing local alignment entirely but keeping the global contrastive loss. We therefore evaluate this adapted method on 18 localized medical tasks on chest X-rays and found that our modified method often achieves similar results to LoVT and in some cases even outperforms it.
%\end{itemize}

% other analysis works
% https://arxiv.org/pdf/2011.02803.pdf
% https://arxiv.org/pdf/2005.10242.pdf
% https://arxiv.org/pdf/2012.09740.pdf
% ViCReg
%repulsive gap: https://arxiv.org/pdf/2203.02053.pdf

% similar methods

\section{Analysis of the Relation between Global and Local Contrastive Losses}
\label{sec:analysis}
We now theoretically study the relationship of the global and local contrastive losses of LoVT~\citep{LoVT}. For a detailed derivation we refer to \cref{sec:derivations}.
Like many image-text contrastive works, LoVT uses two independent encoders (one for images and one for text, i.e.\ radiological reports) to compute global (i.e.\ per-image/per-report) and local (i.e.\ patch/sentence) representations. The global image and text representations (denoted by $\zgs$ and $\zgr$) are then aligned using a NTXent-based~\citep{SimCLR} contrastive loss, denoted by $\mathcal{L}_\text{global}$.
The local representations are first transferred to the other modality using an attention model, such that for each patch representation $\zs$ (of patch $k$ from the image model) we have a representation $\zrs$ with information from the report and vice-versa for each sentence representation $\zr$ (of sentence $m$) we have a representation $\zsr$ with information from the image. The representation pairs $(\zs,\zrs)$ and $(\zr,\zsr)$ are then aligned using the local NTXent-based losses $\mathcal{L}_\text{local-image}$ and $\mathcal{L}_\text{local-report}$, respectively.
Following \citet{uniformity}, each of the three contrastive losses can be decomposed into an \emph{alignment} component, maximizing the similarity of related representations (i.e.\ pulling them close to each other), and a \emph{distribution prior} component, imposing a distribution on the representations to prevent them from collapsing to a constant (i.e.\ pushing non-related representations away form each other).
We can therefore decompose the three losses into three alignment components ($\mathcal{L}_\text{global-align}$ for global, $\mathcal{L}^\S_\text{local-align}$ for local image, and $\mathcal{L}^\R_\text{local-align}$ for local text alignment) and three distribution priors ($\mathcal{L}_\text{global-dist}$ as global, $\mathcal{L}^\S_\text{local-dist}$ as local image, and $\mathcal{L}^\R_\text{local-dist}$ as local report prior). 

\subsection{Alignment Components}\label{sec:alignment}
%We no study the relation of the three alignment components $\mathcal{L}_\text{global-align}$, $\mathcal{L}^\S_\text{local-align}$, and $\mathcal{L}^\R_\text{local-align}$. %All of these components, maximize the cosine similarity between related representations in order to pull them closer to each other.
%While $\mathcal{L}_\text{global-align}$ maximize the similarity of global representations from the same sample, $\mathcal{L}^\R_\text{local-align}$ and $\mathcal{L}^\S_\text{local-align}$ maximize the similarity of local representations that have been aligned with each other
%The local alignment components $\mathcal{L}^\S_\text{local-align}$ and $\mathcal{L}^\R_\text{local-align}$ on the other hand align local representations of image regions or report sentences with local cross-modality representations computed using attention on local representations from the other modality. The individual (local) cosine similarities are aggregated as weighted sums per modality and then maximized.
In order to study the relations between the alignment components $\mathcal{L}_\text{global-align}$, $\mathcal{L}^\S_\text{local-align}$, and $\mathcal{L}^\R_\text{local-align}$, we make some simplifying assumptions: i) We assume that the representations $\zgs$ and $\zsr$ are computed as weighted sums from region representations $\zs$, and similarly $\zgr$ and $\zrs$ as weighted sums from sentence representations $\zr$, and ii) we ignore the normalization terms of the cosine similarity, i.e.\ we treat all cosine similarities as dot products, denoted by $\text{dot}(\cdot, \cdot)$. 
Using these assumptions, and given the batch size $N$, with $K$ patches per image, and $M_i$ sentences in sample $i$, each of the three alignment components can be written as a weighted sum (with weights $\xi_{i, k, m}$) of dot products between region representations $\zs$ and sentence representations $\zr$:
\begin{align}
    \mathcal{L}_\text{align} &= - \frac{1}{N}\sum_{i=1}^N \sum_{k=1}^{K}\sum_{m = 1}^{M_i} \xi_{i, k, m}\text{dot}\left(\zs, \zr\right) \,,
\label{eq:align_rewritten}    
\end{align}
where the exact form of weights $\xi_{i, k, m}$ depends on the specific loss function (see \cref{sec:deriv_alignment} for details).
%For $\mathcal{L}_\text{global-align}$ we have
%\begin{align}
%    \xi_{i, k, m} &= \xi^{\mathcal{G}}_{i, k, m} = \frac{\gamma\weights[k]\weightr[m]}{\tau} \,,
%\label{eq:xi_g}    
%\end{align}
%where $\gamma$ is the coefficient weighting the loss, $\tau$ is the (global) contrastive temperature, and %$\weights$ and $\weightr$ are the weights for computing $\zgs$ from $\zs$ and $\zgr$ from $\zr$, %respectively.
%For $\mathcal{L}^\S_\text{local-align}$ we have
%\begin{align}
%    \xi_{i, k, m} &= \xi^{\S}_{i, k, m} = \frac{\mu \sum_{l = 1}^{K}(\weights \ps + \weights[l] \pslk) %\cdot \alpharsl}{\tau'} \,,
%\label{eq:xi_ls}    
%\end{align}
%where $\mu$ is the coefficient weighting the loss, $\tau'$ is the (local) contrastive temperature, $\ps$ %are computed spatial relationship weights, and $\alpharsl$ are the attention weights for computing $\zsr$ %from $\zs$.
%For $\mathcal{L}^\R_\text{local-align}$ we have
%\begin{align}
%    \xi_{i, k, m} &= \xi^{\R}_{i, k, m} = \frac{\nu\weightr \alphasr }{\tau'} \,,
%\label{eq:xi_ls}    
%\end{align}
%where $\nu$ is the coefficient weighting the loss and $\alphasr$ are the attention weights for computing %$\zrs$ from $\zr$.
%We refer to the LoVT paper \citep{LoVT} for a detailed explanation of the used symbols and definitions of %the loss functions.
Therefore, the only difference between the three alignment components is the definition of $\xi_{i,k,m}$. Assuming there are better and worse aligned pairs of local representations in each sample (i.e.\ pair of image and report), the main difference between those losses is how the best aligned pairs are weighted against the worse aligned pairs. This means that in the special case where the local representations of each modality are constant within each sample, they are identical.
One major difference between global and local alignment components is, that for local alignments $\mathcal{L}^\S_\text{local-align}$ and $\mathcal{L}^\R_\text{local-align}$, the $\xi_{i,k,m}$ are not separable into components containing only information from a single modality (image or report). The reason for this is that the attention weights used to compute $\zrs$ and $\zsr$ are not separable in that way, as they are computed from both, image and report, representations. They therefore allow for pairwise interactions between both modalities, which are not possible with the global alignment component $\mathcal{L}_\text{global-align}$ (as $\xi_{i,k,m}$ is separable in that case). 
Summarizing our findings, we identified the following main differences between the global alignment and the local alignment components:
\begin{itemize}
    \item Local alignment allows for more complex pairwise interactions between local representations, not restricted by the separability of the attention weights.
    \item Local and global alignment differ in how they weight well aligned pairs of local representations compared to less aligned pairs. Local alignment can incorporate pairwise distances between $\zs$ and $\zr$ and may thus focus on the best aligned pairs.
    \item Local and global alignment losses differ in where normalization (of the cosine similarity) happens, i.e.\ before or after the summation over local representations. 
    \item Using global and local alignment together allows for decoupling of local from global representations by using independent projection heads.
\end{itemize}
Despite these differences, we argue that local and global alignment losses both enforce well aligned local as well as global representations. In \cref{sec:experiments} we therefore empirically study removing the local while keeping only the global alignment components.

\subsection{Distribution Prior Components and Uniformity}\label{sec:distribution}
Studying the distribution prior components $\mathcal{L}_\text{global-dist}$, $\mathcal{L}^\S_\text{local-dist}$, and $\mathcal{L}^\R_\text{local-dist}$, we found that they all follow a similar form: They all minimize the (weighted) average of \texttt{logsumexp}-aggregations over pairwise cosine-similarities between weighted sums of the local representations $\zs$ and $\zr$ (see \cref{sec:deriv_distribution}).
However, while in the global distribution prior $\mathcal{L}_\text{global-dist}$ the \texttt{logsumexp}-aggregation is done over the samples in the batch, in the local distribution priors $\mathcal{L}^\S_\text{local-dist}$ and $\mathcal{L}^\R_\text{local-dist}$ the \texttt{logsumexp}-aggregation is done over the local representations of each sample (i.e. we sum over the regions or sentences per sample).
This means that while the $\mathcal{L}_\text{global-dist}$ loss imposes representations to be roughly uniformly distributed over the whole dataset (i.e.\ pushing representations from different samples away from each other), $\mathcal{L}^\S_\text{local-dist}$ and $\mathcal{L}^\R_\text{local-dist}$ impose uniform distributions of (local) representations over each sample (i.e.\ pushing representations within each sample away from other representations from the same sample). 

We argue that the contrast between local representations (including patch representations), which is enforced by the local distributions priors, is essential for the success of pretraining methods on localized downstream tasks. Therefore, instead of removing these local distribution priors, we propose to replace them by distribution priors that impose a per-sample uniform distribution on each modality (image and text) independently.
Following \citet{uniformity}, we use a loss based on pairwise Gaussian potentials but adapt it to impose uniform distributions within each sample instead of imposing them over the whole dataset. This \emph{Gaussian uniformity (uni-gauss)} loss is then applied to the image and report modality independently. For images it is defined as 
\begin{align}
    \mathcal{L}^\S_{\text{uni-gauss}}  & = \frac{1}{N}\sum_{i=1}^N \log \frac{1}{K^2} \sum_{k=1}^K\sum_{k'=1}^K \exp\left(\frac{\cos(\zs, \zs[k'])}{\tau'}\right) \,,
\end{align}
while for reports it is defined as
\begin{align}
    \mathcal{L}^\R_{\text{uni-gauss}} = \frac{1}{N}\sum_{i=1}^N \log \frac{1}{M_i^2} \sum_{m=1}^{M_i}\sum_{m'=1}^{M_i} \exp\left(\frac{\cos(\zr, \zr[m'])}{\tau'}\right) \,.
\end{align}
We will empirically study the effect of this loss in the next section.

\section{Empirical Study}
\label{sec:experiments}
\subsection{Experimental Setup}
Following the insights from \cref{sec:analysis}, we empirically study the effect of removing local alignment components and replacing local distribution components of LoVT's loss function, resulting in:
\begin{align}
\mathcal{L}_\text{LoVT-uni-gauss} &= \gamma \cdot \mathcal{L}_\text{global} + \eta\cdot \left( \mathcal{L}^\S_{\text{uni-gauss}} + \mathcal{L}^\R_{\text{uni-gauss}}\right) \,,
\end{align}
where $\gamma$ (we used the same value as LoVT) and $\eta$ (determined by hyperparameter tuning) are loss coefficients.
Apart from the changes to the loss function, we follow the same framework and training procedure as used in LoVT. Note however that we share the projection heads for local and global representations (per modality). We pretrain on 30\% of MIMIC-CXRv2~\citep{MIMIC-CXR,MIMIC-CXR-2,MIMIC-CXR-JPG} and then evaluate our models trained on the same evaluation framework~\citep{LoVT_MICCAI} as used in LoVT, which consists of 18 localized medical tasks (semantic segmentation and object detection) on five public chest X-ray datasets. 
We compare the results against methods only having global losses, i.e.\ CLIP~\citep{CLIP}, ConVIRT~\citep{ConVIRT}, and LoVT with the local contrastive losses removed (LoVT /wo local), and against the unmodified LoVT.
Note that we experimented with different temperatures $\tau'$ and also studied a cross-entropy-based variant of the proposed uniformity loss, which we call \emph{uni-xent} (see \cref{sec:xent_unif_def}), but found the Gaussian uniformity loss to be more effective. We refer to \cref{sec:ablations} for detailed ablation studies.

\subsection{Results}
\begin{table}[t]
  \caption{Correlation of downstream results of our uniformity-based modification of LoVT and the unmodified LoVT. We consider (in the rows) the number of tasks where uniformity-based models i) outperform CLIP, ii) outperform LoVT, iii) outperform LoVT /wo local, and iv) outperform all studied methods, and (in the columns) on how many of them a) LoVT outperforms CLIP ($\uparrow$) or vice-versa ($\downarrow$) and b) LoVT /wo local performs worse ($\downarrow$) or better ($\uparrow$) than LoVT with local losses.}
  \label{tab:comparison}
  \centering
  %\tiny
  %\setlength{\tabcolsep}{1.5pt}
\input{figures/comparison}

 \end{table} 
 We found that the best LoVT-uni-gauss method outperforms LoVT /wo local on 10 of 18 tasks and CLIP on 9 tasks. On 6 tasks it could even outperform LoVT, showing that LoVT-uni-gauss is competitive with LoVT.
 If we choose the local temperature $\tau'$ individually per downstream task (out of two studied temperatures), the LoVT-uni-gauss method outperforms LoVT /wo local on 12, CLIP on 13 and LoVT on 7 of 18 tasks. This highlights the importance of the local temperature $\tau'$ and indicates that different tasks require different strengths of uniformity. 
 
 In order to study the effect of local uniformity in general, without the restriction to a single uniformity loss variant, we now consider both uniformity variants, uni-gauss and uni-xent. On 12 of 18 tasks (on 10 tasks by more then the 95\%-confidence interval) uniformity based methods
(uni-gauss or uni-xent) outperform all studied image-text methods without local losses (i.e. ConVIRT, CLIP, and LoVT /wo local).
These results indicate that with well-tuned hyperparameters, uniformity-based methods outperform image-text methods that only rely on global losses on the large majority of studied tasks, even significantly on most of them. However, on 9 of 18 tasks LoVT outperforms all uniformity-based methods, from which we conclude that on some tasks the additional local alignment components of LoVT are still beneficial.

We also study how good performance of our uniformity-based methods correlate with good performance of LoVT when comparing them against methods without local losses. In \cref{tab:comparison} we therefore consider the cases (i.e.\ tasks) where uniformity-based methods (uni-gauss or uni-xent) are i) better than CLIP, ii) better than LoVT, iii) better than LoVT /wo local, and iv) better than all studied methods. For each of these cases we then compare on how many of them a) LoVT or CLIP is better and b) LoVT /wo local is better or worse than LoVT.
We observe that on tasks where uniformity-based methods are better than CLIP or better than LoVT /wo local, LoVT often also outperforms CLIP and LoVT /wo local, indicating that local uniformity plays an important role in the success of LoVT. For detailed results we refer to \cref{sec:detailed_results} and for a comparison of the effective uniformity of representations we refer to \cref{sec:uniformity_comparison}.

 %We also realize that uniformity-based methods more often outperform LoVT on tasks where LoVT is outperformed by CLIP. 
 % RSNA Finetune 100, RSNA Frozen 100, Object CXR Finetune, Object CXR Linear.
 %But LoVT is also outperformed on some tasks where LoVT outperforms CLIP, i.e. on COVID Finetune and Frozen, SIIm-ACR Frozen, Object CXR Frozen
%Using the Linear evaluation scheme LoVT performs best (better than uniformity-based methods) on most tasks indicating that for this type of evaluation the alignment loss from LoVT is beneficial. 
%On the COVID dataset (except for Linear evaluation) and on the Object CXR dataset uniformity-based methods typically outperform LoVT. In both cases no or only limited supervision relevant for these tasks was included in the reports of the pretraining dataset. Uniformity-based methods are thus more effective for pretraining with limited text supervision, as they induce per-sample uniform distribution of representations without utilizing the alignment of images and reports.
%On some finetuning tasks, neither LoVT nor uniformity-based methods perform well, whereas image-only methods or classification pretraining are useful. 

%COVID Linear seems to benefit from pairwise local alignment.
%Local alignment losses hurt beformance in RSNA Finetune (10 and 100\%), RSNA Frozen 100\%, SIIM-ACR Finetune, Object CXR Linear. 

\subsection{Conclusion}
We studied the relationship of global and local contrastive losses of image-text methods for localized medical downstream tasks. Considering LoVT's loss functions, we found that the alignment components of the global and local losses are highly related and proposed to drop the local alignment components. We found that the distribution priors of the losses act complementary and proposed a local uniformity loss replacing the local distribution priors while keeping the global priors. Empirically, we found that our proposed method typically works well on chest X-ray tasks where local losses improve the results, indicating that local uniformity plays an important role when pretraining for localized tasks. On some tasks however %, especially when restricted to a fixed temperature over all tasks, 
the uniformity-based approaches are still outperformed by LoVT, suggesting that local contrastive losses cannot fully be replaced by local uniformity losses. We hope that our findings inspire future research to further improve pretraining on localized tasks.

%\section{Discussion}
%\subsection{Limitations of our Study}
%Our empirical analysis focuses on the per-sample uniformity. Prior works\cite{uniformity} suggest that using separated global uniformity losses (i.e.\ also decomposing $\mathcal{L}_\text{global}$) can provide additional improvements and insights. Also, we did not study the effect of keeping the local alignment components $\mathcal{L}^\S_\text{local-align}$ and $\mathcal{L}^\R_\text{local-align}$ but tuning them separately. We will leave this to future work.

%\paragraph{Repulsive Gap}
%

% TODO: repulsive gap: https://arxiv.org/pdf/2203.02053.pdf

% repulsive gap is still encouraged by global dist loss
%\subsection{Limitations of our Proposed Method}
%While we showed that explicitly aligning local representations may not be required for pre-training the image encoder, the pre-trained cannot compute aligned representations of image regions and report sentences. However, these might be required for retrieval tasks. Also, it does not provide an alignment probability matrix that might also be useful for downstream tasks.

\bibliographystyle{unsrtnat}
\bibliography{ms}

\appendix
\input{appendix}

\end{document}

%% file: figures/comparison.tex
%auto-ignore
\small
\begin{tabular}{l||cc|cc||c|c}
\toprule
& \multicolumn{2}{c|}{LoVT vs. CLIP} &  \multicolumn{2}{c||}{LoVT /wo local} & Total &\multirow{2}{*}{Total} \\
& LoVT $\uparrow$ & CLIP $\uparrow$  & $\downarrow$ & $\uparrow$ & $>95\%$ conf. & \\
\hline
Uni $>$ CLIP           & \cellcolor{LightCyan}\textbf{\underline{10 (4)}} & \cellcolor{LightCyan}4 (3) & \cellcolor{LightCyan}\textbf{10 (5)} & \cellcolor{LightCyan}\underline{4 (2)} & 10 & \cellcolor{LightCyan}of 14 \\ % (1), (3), 5, 6, 7, 8, 9, | 10, 11, (13), 14, (15), 16, 17
Uni $>$ LoVT           & \textbf{5} (1) & \cellcolor{LightCyan}\underline{4 \textbf{(3)}} & \textbf{6} (1) & \cellcolor{LightCyan}\underline{3 \textbf{(2)}} & 6 & of 9 \\ % 3, (5), 6, | (10), (11), 14, 15, 16, 17
Uni $>$ LoVT /wo local & \cellcolor{LightCyan}\textbf{\underline{10} (4)} & \cellcolor{LightCyan}\underline{5 \textbf{(4)}} & \cellcolor{LightCyan}\underline{\textbf{12 (7)}} & \cellcolor{LightCyan}3 (2) & 12 & \cellcolor{LightCyan}of 15 \\ % 1, (3), 4, 5, 6, 7, 8, 9, | 10, (11), (12), 14, 15, 16, 17
Uni is best            & \textbf{\underline{6} (2)} & \underline{3 \textbf{(2)}} & \cellcolor{LightCyan}\textbf{\underline{7} (2)} & 2 \underline{(1)} & 4 & of 9 \\% (5), 6, (8), | (10), (11), 14, (15), 16, 17
\hhline{=======}
\multirow{2}{*}{Total} & of & of & of & of & of & of \\ & \textbf{12 (5)} & 6 (4) & \textbf{13 (7)} & 5 (2) & 18 & 18 \\
\bottomrule
\end{tabular}\\
\scriptsize
\textbf{Note:} The numbers in paranthesis always specify the number of tasks where the column is true for more than the 95\%-confidence interval (of the better model). 
Bold numbers indicate that the cell is the best in its row and block.
Underlined numbers indicate that the cells relative number of tasks (normalized by the column total) is larger than of the other cells in the same row and block. Cells containing a majority of tasks (more than half of the total tasks) in their column  are marked with blue background.
% LoVT $>$ CLIP: (1), 5, 7, 8, 9, (10), (11), 12, (13), (14), (16), 18
% CLIP > LoVT: (2), 3, 4, (6), 15, 17
% /w local $>$ /wo local: 1, 4, 5, 7, 8, 9, (10), (11), 12, (14), (15), (16), (18)
% /wo local $>$ /w local: (2), 3, (6), (13), 17

%% file: appendix.tex
%auto-ignore
\section{Relation to other Image-text Methods}
\label{sec:other_methods}
While our study focuses on LoVT's loss functions, we now shortly explain the relation to other text-supervised methods.
\paragraph{Comparison to ConVIRT, CLIP, and ALIGN}
ConVIRT~\citep{ConVIRT}, CLIP~\citep{CLIP}, and ALIGN~\citep{ALIGN} all follow a similar framework as LoVT but they do not align local representations explicitly, i.e.\ they only optimize $\mathcal{L}_\text{global-align}$ and $\mathcal{L}_\text{global-dist}$ but none of the local loss components. However, all of these methods consider only a single sentence per sample (in the case of reports, they randomly sample a single sentence from it). This means that $\mathcal{L}_\text{global-align}$ effectively aligns local report representations (i.e.\ sentences) with global image representations. However, $\mathcal{L}_\text{global-dist}$ still imposes a uniform distribution over the dataset and there is (unlike in LoVT) no component that enforces (local) representations to be uniformly distributed within each sample. Additionally, only CLIP uses (like LoVT) attention pooling to compute image representations while the other methods use average pooling.

\paragraph{Comparison to local-mi}
In the local-MI~\citep{local_MI} work mutual information is maximized between (randomly sampled) sentence representations and the best matching image region representations. This should therefore have a similar effect as minimizing the local alignment components $\mathcal{L}^\S_\text{local-align}$ and $\mathcal{L}^\R_\text{local-align}$. However, when estimating the mutual information they consider the whole dataset. While it is not explicitly stated whether they also include other (non-matching) region-sentence pairs from the same sample during mutual information estimation, it is probable that their mutual information estimation imposes a distribution prior that is rather similar to $\mathcal{L}_\text{global-dist}$ than to the local distribution priors $\mathcal{L}^\S_\text{local-dist}$ and $\mathcal{L}^\R_\text{local-dist}$. 

\paragraph{Comparison to VirTex and ICMLM}
VirTex~\citep{VirTex} and ICMLM~\citep{ICMLM} follow a different framework. Instead of using contrastive losses for pretraining, they use generative objectives by generating captions for images. While generative objectives may implicitly enforce alignment and distributions priors in order to achieve good generation performance, they are not enforced explicitly. Above all, image region representations are not enforced (at least not explicitly) to be distributed uniformly within each sample. 

\section{Comparison of Uniformity}\label{sec:uniformity_comparison}
%\begin{figure}[t]
%    \centering
%    \includegraphics[width=0.48\textwidth]{figures/std_y_a.pdf}
%    \includegraphics[width=0.48\textwidth]{figures/std_y_b.pdf}
%    \caption{Standard deviation (std) of local and global representations. \textbf{Left:} Image ($\ygs$) and image region ($\ys$) representations. \textbf{Right:} Report ($\ygr$) and report sentence ($\yr$) representations. For local representations we additionally show the mean std per-sample, i.e.\ how different representations are within a sample, and the std of the per-sample centroids.
%    The models were trained on 30\% of frontal MIMIC-CXR and then evaluated on the whole test set.}
%    \label{fig:std}
%\end{figure}
\begin{figure}[t]
    \centering
    \includegraphics[width=.4\textwidth]{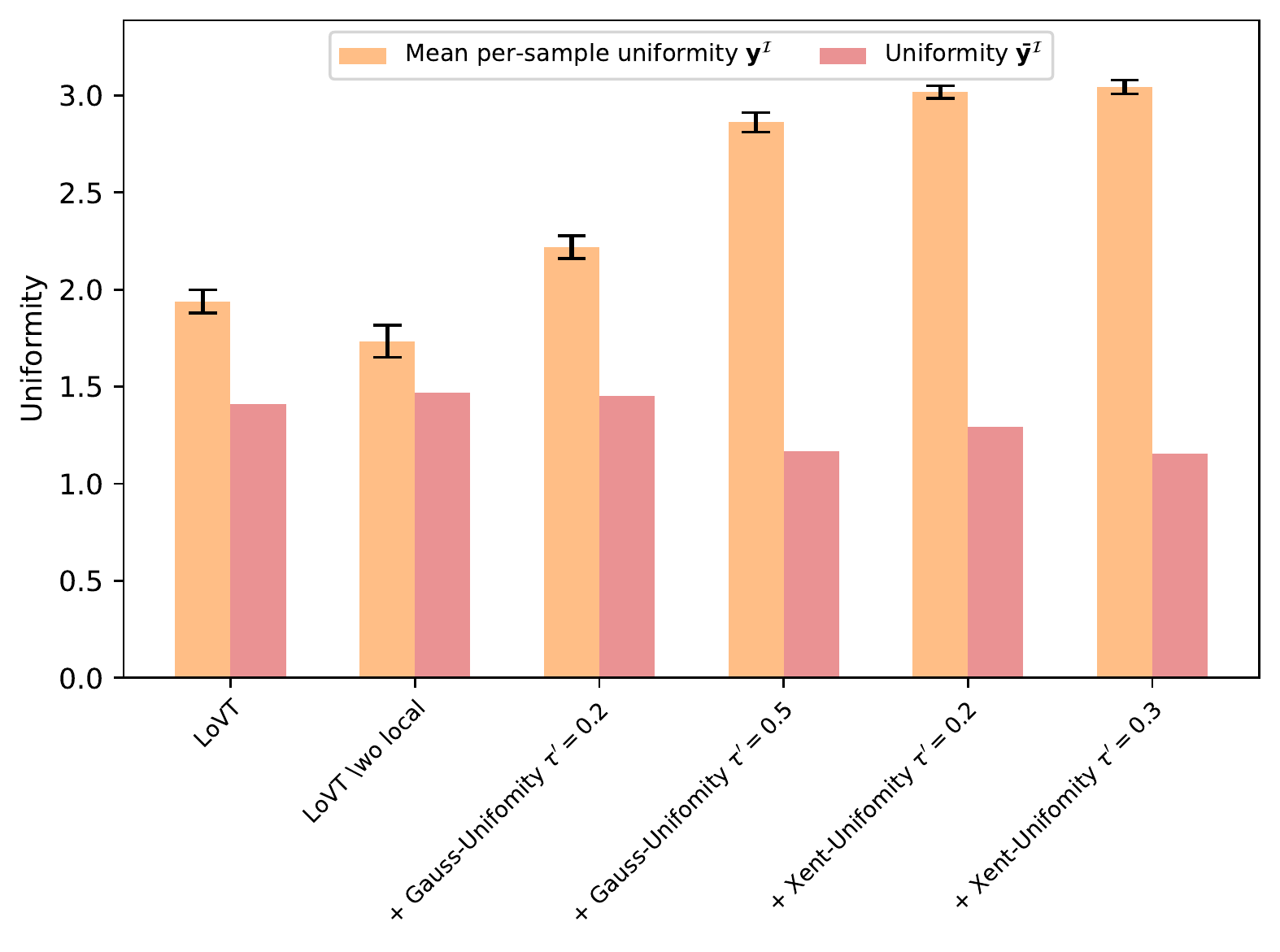}
    \includegraphics[width=.4\textwidth]{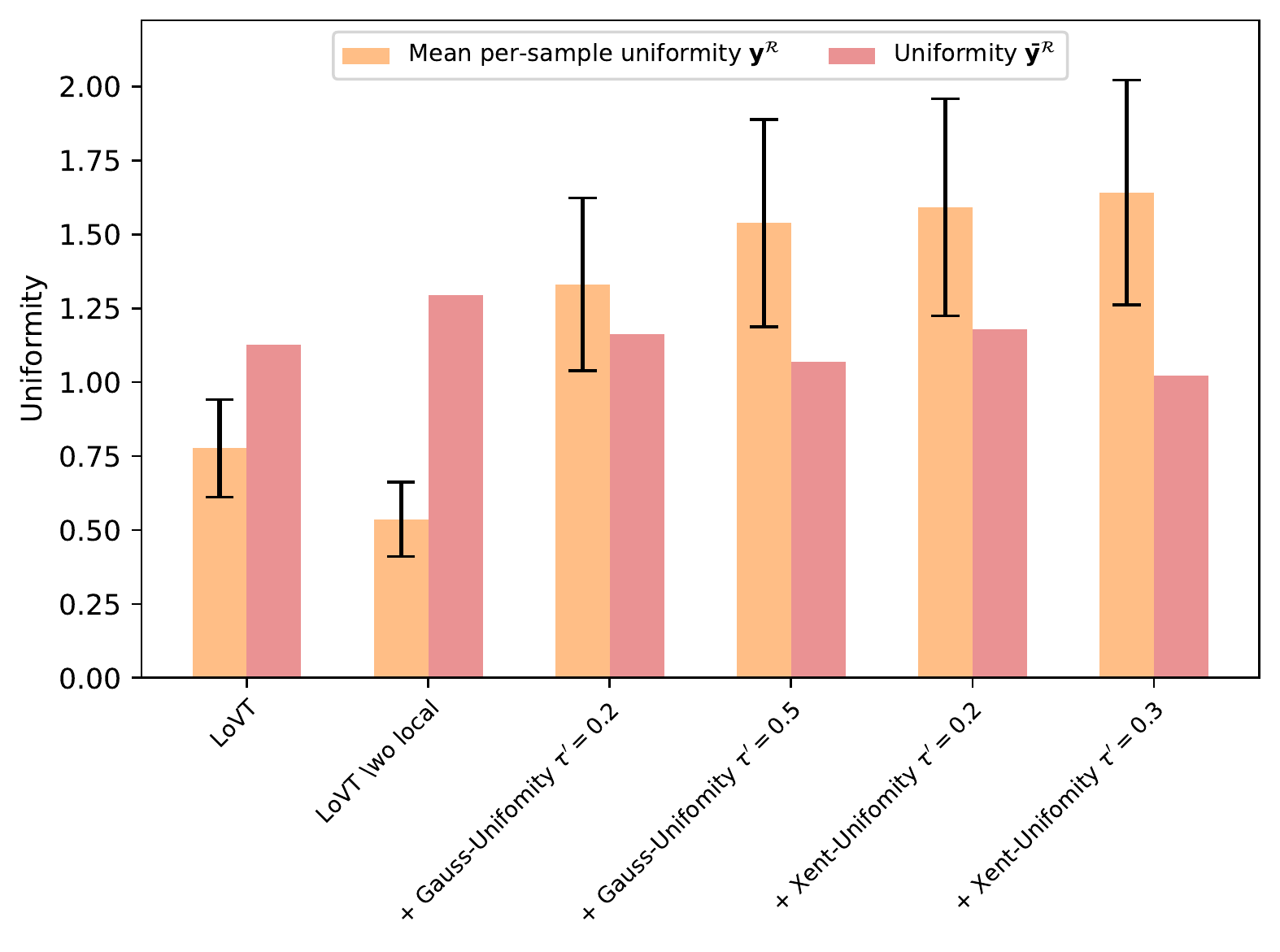}
    \caption{Uniformity of global and per-sample uniformity of local representations. \textbf{Left:} Image ($\ygs$) and image region ($\ys$) representations. \textbf{Right:} Report ($\ygr$) and report sentence ($\yr$) representations.
    The models were trained on 30\% of frontal MIMIC-CXR and then evaluated on the whole test set.}
    \label{fig:uniformity}
\end{figure}
\cref{fig:uniformity} shows the uniformity of local (i.e.\ scan region or report sentence) and global (i.e.\ scan or report) representations of different models. Local uniformity is measured using $-\mathcal{L}^\S_{\text{uni-gauss}}$ and $-\mathcal{L}^\R_{\text{uni-gauss}}$ on the local representations $\ys$ and $\yr$ (before projection), respectively, while the global uniformity is measured using the negative uniformity loss as defined in \citet{uniformity} on the scan $\ygs$ and report $\ygr$ representations (again before projection).
Our local uniformity-based methods have, as expected, much larger per-sample uniformities (especially compared to LoVT /wo local), while at the same time having (slightly) smaller global uniformities.
Larger temperatures increase this effect. Also, the effect is more present in the uni-xent methods compared to the uni-gauss methods.

\FloatBarrier

\section{Detailed Derivations}\label{sec:derivations}
\subsection{Image-text Contrastive Framework of LoVT}
Most image-text contrastive works~\citep{CLIP,ALIGN,ConVIRT,LoVT,local_MI} follow a similar framework. They use two independent encoders, one for encoding images and one for encoding the text (e.g.\ radiological reports). 
In the case of LoVT~\citep{LoVT}, ResNet50~\citep{ResNet50} is used to encode the images and BERT~\citep{BERT} is used to encode the full reports. Each image $\xs$ is encoded into $K = H \times W$ (we use $K = 7 \times 7$) region representations $\ys$, taking the last feature map of the ResNet50. Similarly, each report $\xr$ is encoded into $M_i$ sentence representations $\yr$ by max-pooling over all the token representations (encoded by BERT) of each sentence.
The global (i.e.\ per-sample) representations of images $\ygs$ and reports $\ygr$ are each computed by an attention pooling layer on the region and sentence representations, respectively. All the local and global representations are then projected independently using (non-shared) projection heads based on multi-layer perceptrons (MLPs). We denote the local projected representations by $\zs$ and $\zr$, and the global representations by $\zgs$ and $\zgr$.
LoVT additionally computes cross-modality representations $\zrs$ and $\zsr$ using an attention layer where the local representations of one modality attend to the representations of the other.
These representations are then aligned using LoVT's loss function, which we will recapitulate in the following section.

\subsection{Definition of LoVT's Loss Function}
We will denote LoVT's loss function as $\mathcal{L}_\text{LoVT}$. It consists of three parts, the global contrasive loss $\mathcal{L}_\text{global}$, the local contrastive loss for image regions (aligned with cross-modality sentence representations) $\mathcal{L}_\text{local-image}$ and the local contrastive loss for report sentences (aligned with cross-modality region representations) $\mathcal{L}_\text{local-report}$:
\begin{align}
\mathcal{L}_\text{LoVT} &= \gamma \cdot \mathcal{L}_\text{global} + \mu  \cdot \mathcal{L}_\text{local-image} + \nu  \cdot \mathcal{L}_\text{local-report} \,,
\end{align}
where $\gamma$, $\mu$, and $\nu$ are loss coefficients (hyperparameters).

The global contrastive loss $\mathcal{L}_\text{global}$ is applied to the global representations $\zgs$ and $\zgr$ and is defined as follows:
\begin{align}
    \ell^{\S\|\R}_\text{global} &=- \log \frac{
        e^{\cos\left(\zgs, \zgr\right) / \tau}
    }{
        \sum_{j} e^{\cos\left(\zgs, \zgr[j]\right) / \tau}}
\end{align}
\begin{align}
    \ell^{\R\|\S}_\text{global} &=- \log \frac{
        e^{\cos\left(\zgr, \zgs\right) / \tau}
    }{
        \sum_{j} e^{\cos\left(\zgr, \zgs[j]\right) / \tau}}
\end{align}
\begin{align}
    \mathcal{L}_\text{global} &= \frac{1}{N}\sum_{i=1}^N \left[\lambda\cdot \ell^{\S\|\R}_\text{global} + (1 - \lambda)\cdot \ell^{\R\|\S}_\text{global}\right]\,,
\end{align}
where $\tau$ is the global temperatures, $\lambda \in [0, 1]$ is a hyperparameter, and $N$ denotes the batch size.

The local contrastive loss $\mathcal{L}_\text{local-image}$ for image regions is applied to $\zs$ and $\zrs$ and is defined as:
\begin{align}
    \ell^{\S\|\R\rightarrow \S}_\text{local-image} = -\sum_{l=1}^K \ps \log \frac{
        e^{\cos\left(\zs, \zrs[l]\right) / \tau'}
    }{
        \sum_{k'} e^{\cos\left(\zs, \zrs[k']\right) / \tau'}}
\end{align}
\begin{align}
    \ell^{\R\rightarrow \S\|\S}_\text{local-image} = -\sum_{l=1}^K \ps \log \frac{
        e^{\cos\left(\zrs, \zs[l]\right) / \tau'}
    }{
        \sum_{k'} e^{\cos\left(\zrs,  \zs[k']\right) / \tau'}}
\end{align}
\begin{align}
    \mathcal{L}_\text{local-image} &= \frac{1}{2N}\sum_{i=1}^N \sum_{k = 1}^{K} \weights\cdot\left[\ell^{\S\|\R\rightarrow \S}_\text{local-image} + \ell^{\R\rightarrow \S\|\S}_\text{local-image}\right]\,,
\end{align}
where $\tau'$ is the local temperature, $\weights$ is the weight for image region $k$ (computed based on the image of sample $i$) and $\ps$ is the positiveness weight for the region pair $(k, l)$ (computed based on their spatial distance).

The local contrastive loss $\mathcal{L}_\text{local-report}$ for report sentences is applied to $\zr$ and $\zsr$ and is defined as:
\begin{align}
    \ell^{\R\|\S\rightarrow \R}_\text{local-report} = -\log \frac{
        e^{\cos\left(\zr, \zsr[m]\right) / \tau'}
    }{
        \sum_{m'} e^{\cos\left(\zr, \zsr[m']\right) / \tau'}}
\end{align}
\begin{align}
    \ell^{\S\rightarrow \R\|\R}_\text{local-report} = -\log \frac{
        e^{\cos\left(\zsr, \zr[m]\right) / \tau'}
    }{
        \sum_{m'} e^{\cos\left(\zsr,  \zr[m']\right) / \tau'}}
\end{align}
\begin{align}
    \mathcal{L}_\text{local-report} &= \frac{1}{2N}\sum_{i=1}^N \sum_{m = 1}^{M_i} \weightr\cdot\left[\ell^{\R\|\S\rightarrow \R}_\text{local-report} + \ell^{\S\rightarrow \R\|\R}_\text{local-report}\right]\,,
\end{align}
where $\weightr$ is the weight for sentence $m$ (computed based on the report of sample $i$).

For a detailed explanation of the components and their intuition we refer to the original publication of LoVT~\citep{LoVT}.

\subsection{Decomposition of the Contrastive Losses of LoVT}
\label{sec:derivations_decomp}
We now start our analysis with the decomposition of LoVT's loss function.
Each of the three parts of LoVT's loss function (i.e.\ $\mathcal{L}_\text{global}$, $\mathcal{L}_\text{local-image}$, and $\mathcal{L}_\text{local-report}$) are NTXent-based~\citep{SimCLR} contrastive loss functions and following previous works~\citep{uniformity} can therefore each be decomposed into an \emph{alignment} and a \emph{distribution prior} component.
We decompose $\mathcal{L}_\text{global}$:
\begin{align}
    \mathcal{L}_\text{global} = \mathcal{L}_\text{global-align} + \mathcal{L}_\text{global-dist}
\end{align}
into the global alignment component 
\begin{align}
    \mathcal{L}_\text{global-align} = - \frac{1}{\tau}\frac{1}{N}\sum_{i=1}^N \cos\left(\zgs, \zgr\right)
\end{align}  
and the global distribution prior
\begin{align}
\label{eq:global_dist}
\begin{split}
    \mathcal{L}_\text{global-dist} &= \lambda\frac{1}{N}\sum_{i=1}^N  \log\sum_{j=1}^N \exp\left(\frac{\cos\left(\zgs, \zgr[j]\right)}{\tau}\right) \\
    & \qquad + (1 - \lambda) \frac{1}{N}\sum_{i=1}^N  \log\sum_{j=1}^N \exp\left(\frac{\cos\left(\zgr, \zgs[j]\right)}{\tau}\right)\\
\end{split} 
\end{align}

Similarly, we can decompose the local losses $\mathcal{L}_\text{local-image}$ and $\mathcal{L}_\text{local-report}$:
\begin{align}
    \mathcal{L}_\text{local-image} &= \mathcal{L}^\S_\text{local-align} + \mathcal{L}^\S_\text{local-dist} \\
    \cdot \mathcal{L}_\text{local-report} &= \mathcal{L}^\R_\text{local-align} + \mathcal{L}^\R_\text{local-dist}
\end{align}
into the local alignment component for scan regions
\begin{align}
\begin{split}
    \mathcal{L}^\S_\text{local-align} &= 
    %- \frac{\mu}{\tau'}\frac{1}{2N}\sum_{i=1}^N \sum_{k = 1}^{K} \sum_{l = 1}^{K} \weights \ps \left[\cos\left(\zs, \zrs[l]\right) + \cos\left(\zrs, \zs[l]\right) \right] \\
    %\begin{split}
    %&= - \frac{\mu}{\tau'}\frac{1}{2N}\sum_{i=1}^N \sum_{k=1}^{K}\sum_{l = 1}^{K} \weights \ps \cos\left(\zs, \zrs[l]\right) \\
    %& \quad - \frac{\mu}{\tau'}\frac{1}{2N}\sum_{i=1}^N \sum_{k=1}^{K}\sum_{l = 1}^{K}\weights[l] \pslk  \cos\left(\zs, \zrs[l]\right)
    %\end{split} \\
    %&= 
    - \frac{1}{\tau'}\frac{1}{N}\sum_{i=1}^N \sum_{k=1}^{K}\sum_{l = 1}^{K}\frac{\weights \ps + \weights[l] \pslk}{2} \cos\left(\zs, \zrs[l]\right)
\end{split}
\end{align}
and the alignment for report sentences
\begin{align}
    \mathcal{L}^\R_\text{local-align} = - \frac{1}{\tau'}\frac{1}{N}\sum_{i=1}^N \sum_{m = 1}^{M_i} \weightr \cos\left(\zr, \zsr[m]\right)
\end{align}
as well as into the distribution prior for scan regions
\begin{align}
\label{eq:local_scan_dist}
\begin{split}
    \mathcal{L}^\S_\text{local-dist} &= \frac{1}{2N}\sum_{i=1}^N \sum_{k=1}^{K} \weights \log\sum_{k'=1}^K \exp\left(\frac{\cos\left(\zs, \zrs[k']\right)}{\tau'}\right) \\
    & \qquad + \frac{1}{2N}\sum_{i=1}^N \sum_{k=1}^{K} \weights \log\sum_{k'=1}^K \exp\left(\frac{\cos\left(\zrs, \zs[k']\right)}{\tau'}\right)\\
\end{split} 
\end{align}
and the prior for report sentences
\begin{align}
\label{eq:local_report_dist}
\begin{split}
    \mathcal{L}^\R_\text{local-dist} &= \frac{1}{2N}\sum_{i=1}^N \sum_{m=1}^{M_i} \weightr \log\sum_{m'=1}^{M_i} \exp\left(\frac{\cos\left(\zr, \zsr[m']\right)}{\tau'}\right) \\
    & \qquad + \frac{1}{2N}\sum_{i=1}^N \sum_{m=1}^{M_i} \weightr \log\sum_{m'=1}^{M_i} \exp\left(\frac{\cos\left(\zsr, \zr[m']\right)}{\tau'}\right)\,.
\end{split} 
\end{align}
In the following sections we will now analyze each of these components individually.

\subsection{Alignment Components}\label{sec:deriv_alignment}
We now study the three alignment components $\mathcal{L}_\text{global-align}$, $\mathcal{L}^\S_\text{local-align}$, and $\mathcal{L}^\R_\text{local-align}$.
The global alignment component $\mathcal{L}_\text{global-align}$ forces (global) image and report representations from the same sample to be aligned by maximizing their cosine similarity.
The local alignment components $\mathcal{L}^\S_\text{local-align}$ and $\mathcal{L}^\R_\text{local-align}$ on the other hand align local representations of image regions or report sentences with local cross-modality representations computed using attention on local representations from the other modality. The individual (local) cosine similarities are aggregated as weighted sums per modality and then maximized.

We will now study how the global alignment component is related to the local alignment components and will therefore make some simplifying assumptions. First, we simplify the attention pooling operation used to compute global representations and replace it by a weighted sum (ignoring that attention pooling uses multiple heads and linear projections). Second, we ignore that local representations $\zs$ and $\zr$ are projected independently from the global representations $\zgs$ and $\zgr$.
Instead, we assume that $\zgs$ and $\zgr$ are computed as a weighted sum of local representations:
\begin{align}
\label{eq:zg}
    \zgs = \sum_{k=1}^K \weights[k] \zs \qquad \zgr = \sum_{m=1}^{M_i} \weightr[m] \zr
\end{align}
Similarly, we simplify the computation of the cross-modality representations $\zsr$ and $\zrs$ by ignoring the linear projections of the attention mechanism used in LoVT and assume:
\begin{align}
\label{eq:zcross}
    \zsr = \sum_{k=1}^{K} \alphasr \zs \qquad \zrs = \sum_{m=1}^{M_i} \alphars \zr
\end{align}
We also ignore the normalization terms of the cosine similarity, i.e.\ we treat all cosine similarities as dot products. Note that this is equal to assuming that $\zs$, $\zr$, $\zgs$, $\zgr$, $\zsr$ and $\zrs$ are unit vectors which however does not hold considering \cref{eq:zg,eq:zcross}.

Using these simplifications we can rewrite the alignment losses. The global alignment loss is rewritten as
\begin{align}
    \mathcal{L}_\text{global-align} &= - \frac{1}{\tau}\frac{1}{N}\sum_{i=1}^N \text{dot}\left(\sum_{k=1}^K \weights[k] \zs, \sum_{m=1}^{M_i} \weightr[m] \zr\right) \label{eq:global_align_rewritten_1}
    \\
    &= - \frac{1}{N}\sum_{i=1}^N \sum_{k=1}^K \sum_{m=1}^{M_i} \frac{\weights[k]\weightr[m]}{\tau} \text{dot}\left(\zs, \zr \right) \\
    &= - \frac{1}{N}\sum_{i=1}^N \sum_{k=1}^{K}\sum_{m = 1}^{M_i} \xi^{\mathcal{G}}_{i, k, m}\text{dot}\left(\zs, \zr\right) \,,
\label{eq:global_align_rewritten}    
\end{align}
the local scan alignment as
\begin{align}
    \mathcal{L}^\S_\text{local-align} &= - \frac{1}{\tau'}\frac{1}{N}\sum_{i=1}^N \sum_{k=1}^{K}\sum_{l = 1}^{K}\frac{\weights \ps + \weights[l] \pslk}{2} \text{dot}\left(\zs, \sum_{m=1}^{M_i}\alpharsl \zr\right) \\
    &= - \frac{1}{N}\sum_{i=1}^N \sum_{k=1}^{K}\sum_{m = 1}^{M_i} \frac{\sum_{l = 1}^{K}(\weights \ps + \weights[l] \pslk) \cdot \alpharsl}{\tau'} \text{dot}\left(\zs, \zr\right) \\
    &= - \frac{1}{N}\sum_{i=1}^N \sum_{k=1}^{K}\sum_{m = 1}^{M_i} \xi^{\S}_{i, k, m}\text{dot}\left(\zs, \zr\right) \,,
\label{eq:local_scan_align_rewritten}   
\end{align}
and the local report alignment is rewritten as
\begin{align}
    \mathcal{L}^\R_\text{local-align} &= - \frac{1}{\tau'}\frac{1}{N}\sum_{i=1}^N \sum_{m = 1}^{M_i} \weightr \text{dot}\left(\zr, \sum_{m = 1}^{M_i}\alphasr\zs\right) \\
    &= - \frac{1}{N}\sum_{i=1}^N \sum_{k=1}^{K}\sum_{m = 1}^{M_i}\frac{\weightr \alphasr }{\tau'}\text{dot}\left(\zr,\zs\right) \\
    &= - \frac{1}{N}\sum_{i=1}^N \sum_{k=1}^{K}\sum_{m = 1}^{M_i}\xi^{\R}_{i, k, m}\text{dot}\left(\zr,\zs\right) \,.
\label{eq:local_report_align_rewritten}
\end{align}
Comparing \cref{eq:global_align_rewritten,eq:local_scan_align_rewritten,eq:local_report_align_rewritten}, 
we realize that
they are all special cases of the same general form of alignment loss that maximizes a weighted sum of all possible pairs of cosine similarities between two local representations from different modalities. 
The only difference between the three losses is how the weights of the sum, i.e.\ $\xi_{i,k,m}$, are defined. 
%Assuming there are better an worse aligned pairs of local representations in each sample (i.e.\ pair of scan and report), the main difference between those losses is how the best aligned pairs are weighted against the worse aligned pairs. This means that in the special case where the local representations of each modality are the same within each sample, they are identical.

We also realize that $\xi^{\S}_{i, k, m}$ and $\xi^{\R}_{i, k, m}$ are, unlike $\xi^{\mathcal{G}}_{i, k, m}$, not separable into components containing only information from a single modality (image or report). Thus, $\mathcal{L}^\S_\text{local-align}$ and $\mathcal{L}^\R_\text{local-align}$ cannot be rewritten in the original form of $\mathcal{L}_\text{global-align}$.
If we assume that $\alpharsl$ and $\alphasr$ are not computed as in LoVT but are instead separable by modality, i.e.\ $\alpharsl = \tilde{\alpha}^{\R \rightarrow \S}_{i, l}\hat{\alpha}^{\R \rightarrow \S}_{i, m}$ and $\alphasr = \tilde{\alpha}^{\S \rightarrow \R}_{i, m}\hat{\alpha}^{\S \rightarrow \R}_{i, k}$, then we can rewrite $\xi^{\S}_{i, k, m}$ and $\xi^{\R}_{i, k, m}$ as
\begin{align}
\xi^{\S}_{i, k, m} &= \frac{1}{\tau'}\sum_{l = 1}^{K}(\weights \ps + \weights[l] \pslk) \cdot \tilde{\alpha}^{\R \rightarrow \S}_{i, l}\hat{\alpha}^{\R \rightarrow \S}_{i, m} \\
\xi^{\R}_{i, k, m} &= \frac{1}{\tau'}\weightr \tilde{\alpha}^{\S \rightarrow \R}_{i, m}\hat{\alpha}^{\S \rightarrow \R}_{i, k}
\end{align}
%\sqrt{\frac{\gamma}{\tau}}\weights \cdot \sqrt{\frac{\gamma}{\tau}} \weightr + 
and therefore as
\begin{align}
\begin{split}
 \xi^{\S}_{i, k, m} = \tilde{\xi}^{\S}_{i, k}\hat{\xi}^{\S}_{i, m} \qquad \xi^{\R}_{i, k, m} = \tilde{\xi}^{\R}_{i, k}\hat{\xi}^{\R}_{i, m}
\end{split}  
\end{align}
with
\begin{align}
 \tilde{\xi}^{\S}_{i, k} &= \sqrt{\frac{1}{\tau'}}\sum_{l = 1}^{K}(\weights \ps + \weights[l] \pslk) \cdot \tilde{\alpha}^{\R \rightarrow \S}_{i, l} &
 \hat{\xi}^{\S}_{i, m} &= \sqrt{\frac{1}{\tau'}}\hat{\alpha}^{\R \rightarrow \S}_{i, m} \\
 \tilde{\xi}^{\R}_{i, k} &= \sqrt{\frac{1}{\tau'}}\hat{\alpha}^{\S \rightarrow \R}_{i, k} &
 \hat{\xi}^{\R}_{i, m} &= \sqrt{\frac{1}{\tau'}}\weightr \tilde{\alpha}^{\S \rightarrow \R}_{i, m}
\end{align}

Using these definitions, we can write 
\begin{align}
\label{eq:local_scan_align_separable}
\begin{split}
    \mathcal{L}^\S_\text{local-align} &= - \frac{1}{N}\sum_{i=1}^N \sum_{k=1}^{K}\sum_{m = 1}^{M_i} \tilde{\xi}^{\S}_{i, k}\hat{\xi}^{\S}_{i, m}\text{dot}\left(\zs, \zr\right) \\
    &= - \frac{1}{N}\sum_{i=1}^N \text{dot}\left(\sum_{k=1}^{K}\tilde{\xi}^{\S}_{i, k}\zs, \sum_{m = 1}^{M_i}\hat{\xi}^{\S}_{i, m}\zr\right)
\end{split}  
\end{align}
and
\begin{align}
\label{eq:local_report_align_separable}
\begin{split}
    \mathcal{L}^\R_\text{local-align} &= - \frac{1}{N}\sum_{i=1}^N \sum_{k=1}^{K}\sum_{m = 1}^{M_i} \tilde{\xi}^{\R}_{i, k}\hat{\xi}^{\R}_{i, m}\text{dot}\left(\zs, \zr\right) \\
    &= - \frac{1}{N}\sum_{i=1}^N \text{dot}\left(\sum_{k=1}^{K}\tilde{\xi}^{\R}_{i, k}\zs, \sum_{m = 1}^{M_i}\hat{\xi}^{\R}_{i, m}\zr\right)
\end{split}  
\end{align}
Comparing \cref{eq:local_scan_align_separable,eq:local_report_align_separable} with the global alignment loss as defined in \cref{eq:global_align_rewritten_1}, we see that they only differ in the computations of the weights in the weighted sums within the cosine similarity.
We therefore conclude, that the non-separability of $\xi^{\S}_{i, k, m}$ and $\xi^{\R}_{i, k, m}$ is an important aspect when comparing the global with the local contrastive losses.

\subsection{Distribution Prior Components}\label{sec:deriv_distribution}
\subsubsection{Comparison of the Distributions Priors}
We now consider the distribution priors of the loss function decomposition.
The global distribution component $\mathcal{L}_\text{global-dist}$ imposes a prior on the distribution of image and sentence representations such that they do not collapse to a constant value. 
Minimizing $\mathcal{L}_\text{global-dist}$ introduces contrast by pushing representations from both modalities away from each other, i.e. maximizing the distance between $\zgs$ and $\zgr[j]$ for all $i, j$. Considering that $\mathcal{L}_\text{global-align}$ pushes the scan and report representations from the same sample $i$ closer to each other, it can be expected that $\mathcal{L}_\text{global-dist}$ pushes representations from different samples ($i \neq j$) further apart than the representations from the same sample ($i=j$). It thus pushes representations (even from the same modality) further apart and therefore imposes a uniform distribution on the representations from each modality, i.e.\ it enforces both, $\zgs$ and $\zgr$, to be roughly uniformly distributed. To understand this, consider the extreme case where $\mathcal{L}_\text{global-align}$ is constant, i.e. $\cos(\zgs, \zgr) = c_i$ for each $i$. The only way to minimize $\mathcal{L}_\text{global-dist}$ is then to also push representations from the same modality (e.g.\ $\zgs[j]$ for different samples $j \neq i$) further apart, i.e.\ to maximize $\cos(\zgs, \zgs[j])$ for $i \neq j$.

For comparing the global distribution prior $\mathcal{L}_\text{global-dist}$ to the local distribution priors $\mathcal{L}^\S_\text{local-dist}$ and $\mathcal{L}^\R_\text{local-dist}$, we first rewrite them by again assuming \cref{eq:zg,eq:zcross} hold, such that we have
\begin{align}
\begin{split}
    \mathcal{L}_\text{global-dist} &= \lambda\frac{1}{N}\sum_{i=1}^N  \log\sum_{j=1}^N \exp\left(\frac{\cos\left(\sum_{k=1}^K \weights[k] \zs, \sum_{m=1}^{M_i} w^{\R}_{j, m} \zrj\right)}{\tau}\right) \\
    & \qquad + (1 - \lambda) \frac{1}{N}\sum_{i=1}^N  \log\sum_{j=1}^N \exp\left(\frac{\cos\left(\sum_{m=1}^{M_i}\weightr\zr, \sum_{k=1}^Kw^{\S}_{j, s} \zsj\right)}{\tau}\right) \,.
\end{split} 
\label{eq:global_dist_rewritten}
\end{align}
Similarly, we rewrite $\mathcal{L}^\S_\text{local-dist}$ as
\begin{align}
\label{eq:local_scan_dist_rewritten}
\begin{split}
    \mathcal{L}^\S_\text{local-dist} &= \frac{1}{2N}\sum_{i=1}^N \sum_{k=1}^{K} \weights \log\sum_{k'=1}^K \exp\left(\frac{\cos\left( \zs, \sum_{m=1}^{M_i} \alphars[k']\zr\right)}{\tau'}\right) \\
    & \qquad + \frac{1}{2N}\sum_{i=1}^N \sum_{k=1}^{K} \weights \log\sum_{k'=1}^K \exp\left(\frac{\cos\left(\sum_{m=1}^{M_i} \alphars\zr, \zs[k']\right)}{\tau'}\right)
\end{split} 
\end{align}
and $\mathcal{L}^\R_\text{local-dist}$ as
\begin{align}
\begin{split}
    \mathcal{L}^\R_\text{local-dist} &= \frac{1}{2N}\sum_{i=1}^N \sum_{m=1}^{M_i} \weightr \log\sum_{m'=1}^{M_i} \exp\left(\frac{\cos\left(\zr, \sum_{k=1}^{K} \alphasr[m'] \zs\right)}{\tau'}\right) \\
    & \qquad + \frac{1}{2N}\sum_{i=1}^N \sum_{m=1}^{M_i} \weightr \log\sum_{m'=1}^{M_i} \exp\left(\frac{\cos\left(\sum_{k=1}^{K} \alphasr \zs, \zr[m']\right)}{\tau'}\right) \,.
\end{split} 
\label{eq:local_report_dist_rewritten}
\end{align}
Comparing \cref{eq:global_dist_rewritten} with \cref{eq:local_scan_dist_rewritten} and \cref{eq:local_report_dist_rewritten}, we realize that they follow a similar form as they both minimize the (weighted) average of \texttt{logsumexp}-aggregations over pairwise cosine-similarities between local representations $\zs$ and $\zr$ (or weighted sums of them). However while in $\mathcal{L}_\text{global-dist}$ the  \texttt{logsumexp}-aggregation is done over the samples in the batch (i.e.\ we sum over $N$ samples), in $\mathcal{L}^\S_\text{local-dist}$ and $\mathcal{L}^\R_\text{local-dist}$ \texttt{logsumexp}-aggregation is done over the local representations of each sample (i.e.\ we sum over $K$ or $M_i$ regions or sentences, respectively, per sample).
This means that while the $\mathcal{L}_\text{global-dist}$ loss imposes representations to be roughly uniformly distributed over the whole dataset, $\mathcal{L}^\S_\text{local-dist}$ and $\mathcal{L}^\R_\text{local-dist}$ impose uniform distributions of (local) representations over each sample, i.e.\ it pushes (local) representations within each sample away from each other and not across samples. 

\subsubsection{Cross-Entropy Uniformity Applied to each Modality Independently}\label{sec:xent_unif_def}
We now propose to replace the local distribution component $\mathcal{L}^\S_\text{local-dist}$ and $\mathcal{L}^\R_\text{local-dist}$ by distribution components that impose a per-sample uniform distribution on each modality independently. Therefore, instead of using cosine similarities between (local) representations from different modalities, we apply the cosine similarities to pairs of (local) representations from the same modality. Additionally, we ignore the local weights $\weights$ and $\weightr$ and use unweighted averages instead. We call the resulting local distributions components \emph{cross-entropy uniformity (uni-xent)} as it is similar to the distribution component of the cross-entropy loss. We define the (local) cross-entropy uniformity for scans as:
\begin{align}
\mathcal{L}^\S_{\text{uni-xent}} & = \frac{1}{N}\sum_{i=1}^N \frac{1}{K}\sum_{k=1}^K \log \sum_{k'=1}^K \exp\left(\frac{\cos(\zs, \zs[k'])}{\tau'}\right)
\end{align}
and for reports as:
\begin{align}
\mathcal{L}^\R_{\text{uni-xent}} & = \frac{1}{N}\sum_{i=1}^N \frac{1}{M_i}\sum_{m=1}^{M_i} \log \sum_{m'=1}^{M_i} \exp\left(\frac{\cos(\zr, \zr[m'])}{\tau'}\right) \,.
\end{align}
%where $\eta$ is a hyperparameter that weights the local uniformity loss against the global losses. %Note that we use the same $eta$ for $\mathcal{L}^\S_{\text{uni-xent}}$ and $\mathcal{L}^\R_{\text{uni-xent}}$.

Like the local distribution components $\mathcal{L}^\S_\text{local-dist}$ and $\mathcal{L}^\R_\text{local-dist}$, the uniformity losses $\mathcal{L}^\S_{\text{uni-xent}}$ and $\mathcal{L}^\R_{\text{uni-xent}}$ minimize the cosine similarity of region (or sentence) representations within each sample and therefore push local representations within each sample away from each other. However, unlike $\mathcal{L}^\S_\text{local-dist}$ and $\mathcal{L}^\R_\text{local-dist}$, $\mathcal{L}^\S_{\text{uni-xent}}$ and $\mathcal{L}^\R_{\text{uni-xent}}$ do not have a repulsive effect between the modalities, i.e.\ they do not push regions and sentence representations away from each other. The global distribution prior $\mathcal{L}_\text{global-dist}$ has a similar repulsive effect and we therefore argue that we can replace $\mathcal{L}^\S_\text{local-dist}$ by $\mathcal{L}^\S_{\text{uni-xent}}$ and $\mathcal{L}^\R_\text{local-dist}$ by $\mathcal{L}^\R_{\text{uni-xent}}$ while still imposing similar prior distributions. %We do however not study this effect empirically.

\subsubsection{Pairwise Gaussian Potential}
Following \citet{uniformity} we now study replacing the cross-entropy-based (local) uniformity losses $\mathcal{L}^\S_{\text{uni-xent}}$ and $\mathcal{L}^\R_{\text{uni-xent}}$ by losses based on pairwise Gaussian potentials. We therefore adapt the uniformity loss proposed by \citet{uniformity} to impose uniform distribution within each sample instead of over the whole dataset. This leads to the \emph{Gaussian uniformity (uni-gauss)} loss. For scans we denote it as $\mathcal{L}^\S_{\text{uni-gauss}}$ and define it as follows:
\begin{align}
    \mathcal{L}^\S_{\text{uni-gauss}}  & = \frac{1}{N}\sum_{i=1}^N \log \frac{1}{K^2} \sum_{k=1}^K\sum_{k'=1}^K \exp\left(\frac{-\|\tilde{\bm{z}}^\S_{i, k} - \tilde{\bm{z}}^\S_{i, k'}\|_2^2}{2\tau'}\right) \\
    & = \frac{1}{N}\sum_{i=1}^N \log \frac{1}{K^2} \sum_{k=1}^K\sum_{k'=1}^K \exp\left(\frac{2 \cdot \cos(\zs, \zs[k']) - 2}{2\tau'}\right) \\
    & = \frac{1}{N}\sum_{i=1}^N \log \frac{1}{K^2} \sum_{k=1}^K\sum_{k'=1}^K \exp\left(\frac{\cos(\zs, \zs[k'])}{\tau'}\right) \cdot C \,,
\end{align}
where $C$ is come constant and can therefore be ignored (or integrated into the loss coefficient) when minimizing $\mathcal{L}^\S_{\text{uni-gauss}}$, leading to our final definition:
\begin{align}
    \mathcal{L}^\S_{\text{uni-gauss}}  & = \frac{1}{N}\sum_{i=1}^N \log \frac{1}{K^2} \sum_{k=1}^K\sum_{k'=1}^K \exp\left(\frac{\cos(\zs, \zs[k'])}{\tau'}\right) \,.
\end{align}
Similarly, we define the uniformity loss for reports as
\begin{align}
    \mathcal{L}^\R_{\text{uni-gauss}} = \frac{1}{N}\sum_{i=1}^N \log \frac{1}{M_i^2} \sum_{m=1}^{M_i}\sum_{m'=1}^{M_i} \exp\left(\frac{\cos(\zr, \zr[m'])}{\tau'}\right) \,.
\end{align}
In \cref{sec:uniformity_ablation}, we empirically compare the effects of cross-entropy and Gaussian uniformity.

\section{Ablation Studies and Hyperparameters}\label{sec:ablations}

\subsection{Effects of Temperature and Uniformity Variant}\label{sec:uniformity_ablation}
\begin{table*}[h]
  \centering
  \label{tab:comparison_methods}
  \caption{Comparison of different uniformity-based models (uni-gauss and uni-xent with different local temperatures $\tau'$) on the evaluation tasks. For each model we show on how many of the 18 evaluation tasks it i) outperformed uniformity-based models with the same type of uniformity loss but with different temperatures $\tau'$, ii) other uniformity-based models, iii) LoVT without local losses, iv) LoVT, v) CLIP, and vi) on how many it is the best of all studied methods (including the image-only and supervised baselines studied by \citet{LoVT_MICCAI}). In brackets, we additionally show the number of tasks considering the 95\%-confidence interval over five evaluation runs. For detailed results we refer to \cref{sec:detailed_results}.}
  \tiny
  \setlength{\tabcolsep}{1.5pt}
\input{figures/comparison_methods}

 \end{table*} 

% \begin{table*}[h!]
%  \centering
%  \tiny
%  \setlength{\tabcolsep}{1.5pt}
%\input{figures/comparison_uni}
% \end{table*} 
 
 \FloatBarrier
  \pagebreak
\subsection{Shared Head and Attention Pooling}
\begin{table}[h!]
\centering
\caption{Effect of sharing the projection heads for local representations ($\zs$ and $\zr$, used in the uniformity losses) with the heads for global representations ($\zgs$ and $\zgr$), and the effect of using attention pooling to compute global representations. Note that attention heads are only shared between representations of the same modality, i.e.\ image and report representations are always projected independently. When no attention pooling is used, global representations are instead computed using global average pooling. We found that using shared attention heads is in general beneficial. We assume that this assures a stronger coupling between the alignment effect of the global contrastive loss and the distribution priors of the local uniformity losses. We also found that attention pooling is beneficial, confirming the results of \citet{LoVT}.}
\tiny
\setlength{\tabcolsep}{1.5pt}
\begin{tabular}{lccc>{\columncolor{Gray}}cc}
\toprule
Uniformity & $\tau'$ & Shared head & Att. Pool & \makecell{RSNA \\ YOLOv3 Frozen \\ 10\%} & \makecell{RSNA \\ Lin. Seg. \\ 10\%} \\\midrule
\multirow{3}{*}{Gauss-Uniformity} & \multirow{3}{*}{0.2} & \cellcolor{LightCyan}\cmark & \cellcolor{LightCyan}\cmark & \cellcolor{LightCyan}\textbf{18.4} & \cellcolor{LightCyan}48.5 \\
&& \cmark & \xmark & 14.9 & 49.3 \\
&& \xmark & \cmark & 15.6 & 48.2 \\\midrule
\multirow{3}{*}{Gauss-Uniformity} & \multirow{3}{*}{0.5} & \cellcolor{LightCyan}\cmark & \cellcolor{LightCyan}\cmark & \cellcolor{LightCyan}\textbf{17.6} & \cellcolor{LightCyan}49.4 \\
&& \cmark & \xmark & 15.9 & 49.2 \\
&& \xmark & \cmark & 16.4 & 49.1 \\\midrule
%\multirow{3}{*}{Xent-Uniformity} & \multirow{3}{*}{0.1} & \cellcolor{LightCyan}\cmark & \cellcolor{LightCyan}\cmark & \cellcolor{LightCyan}\textbf{17.2} & \cellcolor{LightCyan}49.4 \\
%&& \cmark & \xmark & 16.0 & 48.1 \\
%&& \xmark & \cmark & 14.3 & 48.5 \\\midrule
\multirow{3}{*}{Xent-Uniformity} & \multirow{3}{*}{0.2} & \cellcolor{LightCyan}\cmark & \cellcolor{LightCyan}\cmark & \cellcolor{LightCyan}\textbf{17.4} & \cellcolor{LightCyan}49.3 \\
&& \cmark & \xmark & 16.3 & 49.2 \\
&& \xmark & \cmark & 16.4 & 49.2 \\\midrule
\multirow{3}{*}{Xent-Uniformity} & \multirow{3}{*}{0.3} & \cellcolor{LightCyan}\cmark & \cellcolor{LightCyan}\cmark & \cellcolor{LightCyan}\textbf{17.6} & \cellcolor{LightCyan}49.2 \\
&& \cmark & \xmark & 15.9 & 48.5 \\
&& \xmark & \cmark & 17.1 & 49.6 \\
\bottomrule
\end{tabular}
\end{table}
  \FloatBarrier
 
\subsection{Hyperparameter Tuning Results}
\begin{table}[h!]
\centering
\caption{Hyperparameter tuning results of the local temperature $\tau'$, tuned using the evaluation task \emph{RSNA YOLOv3 Frozen 10\%}. The coefficient $\eta$ was fixed to $0.25$ for Gauss-Unifomity and to $0.5$ for Xent-Uniformity. We also show the results on \emph{RSNA Lin. Seg. 10\%}. In our main studies we use the two best temperatures per variant (marked in blue).}
\tiny
\setlength{\tabcolsep}{1.5pt}
\begin{tabular}{lc>{\columncolor{Gray}}cc}
\toprule
Uniformity & $\tau'$ & \makecell{RSNA \\ YOLOv3 Frozen \\ 10\%} & \makecell{RSNA \\ Lin. Seg. \\ 10\%} \\\midrule
%No & - & 17.4 & \\\midrule
%No & - & 16.5 & \\\midrule
\multirow{5}{*}{Gauss-Uniformity} & 0.1 & 15.3 & 48.9 \\
 & \cellcolor{LightCyan}0.2 & \cellcolor{LightCyan}\textbf{18.4} & \cellcolor{LightCyan}48.5 \\
& 0.3 & 16.0 & 49.3 \\
& \cellcolor{LightCyan}0.5 & \cellcolor{LightCyan}17.6 & \cellcolor{LightCyan}49.4 \\
& 1.0 & 15.9 & 49.6 \\
\midrule
\multirow{5}{*}{Xent-Uniformity} & 0.05 & 16.1 & 49.1 \\
& 0.1 & 17.2 & 49.4 \\
& \cellcolor{LightCyan}0.2 & \cellcolor{LightCyan}17.4 & \cellcolor{LightCyan}49.3 \\
& \cellcolor{LightCyan}0.3 & \cellcolor{LightCyan}\textbf{17.6} & \cellcolor{LightCyan}49.2 \\
& 0.5 & 14.9 & 49.7 \\
\bottomrule
\end{tabular}
\end{table}

\begin{table}[h!]
\centering
\caption{Hyperparameter tuning results of the uniformity loss coefficient $\eta$, tuned using the evaluation task \emph{RSNA YOLOv3 Frozen 10\%}. We also show the results on \emph{RSNA Lin. Seg. 10\%}. In our main studies we use the configurations marked in blue.}
\tiny
\setlength{\tabcolsep}{1.5pt}
\begin{tabular}{lcc>{\columncolor{Gray}}cc}
\toprule
Uniformity & $\tau'$ & $\eta$ & \makecell{RSNA \\ YOLOv3 Frozen \\ 10\%} & \makecell{RSNA \\ Lin. Seg. \\ 10\%} \\\midrule
\multirow{3}{*}{Gauss-Uniformity} & \multirow{3}{*}{0.2} & 0.1 & 15.5 & 50.0 \\
& & \cellcolor{LightCyan}0.25 & \cellcolor{LightCyan}\textbf{18.4} & \cellcolor{LightCyan}48.5 \\
& & 0.5 & 15.7 & 49.0 \\\midrule
\multirow{3}{*}{Gauss-Uniformity} & \multirow{3}{*}{0.5} & 0.1 & 16.5 & 49.7 \\
& & \cellcolor{LightCyan}0.25 & \cellcolor{LightCyan}\textbf{17.6} & \cellcolor{LightCyan}49.4 \\
& & 0.5 & 16.6 & 49.2 \\\midrule
%\multirow{3}{*}{Xent-Uniformity} & \multirow{3}{*}{0.1} & 0.1 & 14.4 & 49.1 \\
%&& \cellcolor{LightCyan}0.25 & \cellcolor{LightCyan}\textbf{17.2} & \cellcolor{LightCyan}49.4 \\
%&& 0.5 & 15.7 & 49.0 \\\midrule
\multirow{3}{*}{Xent-Uniformity} & \multirow{3}{*}{0.2} & 0.25 & 16.2 & 49.2 \\
&& \cellcolor{LightCyan}0.5 & \cellcolor{LightCyan}\textbf{17.4} & \cellcolor{LightCyan}49.3 \\
&& 0.75 & 15.6 & 49.6 \\\midrule
\multirow{3}{*}{Xent-Uniformity} & \multirow{3}{*}{0.3} & 0.25 & 15.8 & 50.0 \\
&& \cellcolor{LightCyan}0.5 & \cellcolor{LightCyan}\textbf{17.6} & \cellcolor{LightCyan}49.2 \\
&& 0.75 & 13.9 & 49.4 \\
\bottomrule
\end{tabular}
\end{table}
 \FloatBarrier

\section{Detailed Experiment Results}\label{sec:detailed_results}
\begin{table*}[h!]
  \centering
  \caption{Results on the RSNA pneumonia detection tasks with different training set sizes. All results are averaged over five evaluation runs and the 95\%-confidence interval over these runs is shown. The best results per task are underlined, the second best results are dash-underlined and the best results per block are highlighted in bold. Note that the \emph{RSNA YOLOv3 Frozen 10\%} task was used for tuning of all methods and may therefore not be representative as methods may overfit on this task.}
  \tiny
  \setlength{\tabcolsep}{1.5pt}
  \input{figures/results_rsna}\\
  * Modified to use the same image and text encoders as ConVIRT and LoVT.
  \label{tab:results_1}
\end{table*}
\begin{table*}[h!]
  \centering
  \caption{Results on downstream tasks on the COVID Rural, SIIM Pneumothorax, Object CXR, and NIH CXR datasets. All results are averaged over five evaluation runs and the 95\%-confidence interval over these runs is shown. The best results per task are underlined, the second best results are dash-underlined and the best results per block are highlighted in bold.}
  \tiny
  \setlength{\tabcolsep}{1.5pt}
  \input{figures/results_other}\\
  * Modified to use the same image and text encoders as ConVIRT and LoVT.
  \label{tab:results_2}
\end{table*}

%% file: figures/comparison_methods.tex
%auto-ignore
\begin{tabular}{l||cc|ccc|c}
\toprule
& Best $\tau'$ & Best Uni & $>$ /wo local & $>$ LoVT & $>$ CLIP & Best model \\
\hline
\rowcolor{LightCyan}Uni-Gauss $\tau' = 0.2$ & 10 (3) & 6 (1) & 10 (5) & 6 (2) & 9 (5) & 3 (1) \\
% (1), (2), (5), 6, (10), (11), (12), (13), 17, 18 & (1), (2), (5), 6, (10),(12) & >wo 1, (4), 5, 6, (7), (8), 10, (12), (15), 17 & >LoVT (3), (5), 6, (10), (15), 17 & > CLIP (1), 5, 6, (7), 9, (10), 11, (13), 17 & best (5), 6, (10)
Uni-Gauss $\tau' = 0.5$ & 7 (6) & 5 (3) & 11 (9) & 7 (6) & 10 (6) & 3 (2) \\
% 3, (4), 7, 8,  14, 15, 16 & (4), 7, 8, (15), 16 &  >wo (1), 4, 5, (6), 7, 8, 10, 14, 15, 16, 17 & >LoVT 3, 6, (10), 14, 15, 16, 17 & >CLIP 5, 7, 8, 9, (10), (11), (13), 14, (15), 16 & 8, (15), 16
Uni-Gauss any $\tau'$   & -     & 11 (5) & 12 (11) & 8 (6) & 13 (9) & 3 (2) \\
% & 1, (2), (4), (5), 6, 7, 8, (10),(12), (15), 16 & >wo 1, 4, 5, 6, 7, 8, 10, (12), 14, 15, 16, 17 & >LoVT 3, (5), 6, (10), 14, 15, 16, 17 & >CLIP (1), 5, 6, 7, 8, 9, (10), 11, (13), 14, (15), 16, 17 & (10), (15), 16 & best (5), 6, 8
\hline
Uni-Xent $\tau' = 0.2$  & 13 (7) & 5 (1) & 12 (7) & 7 (5) & 12 (7) & 2 (1) \\
% (1), 3, (4), 6, 8, (10), (11), (12), (13), 15, 16, 17, 18 & (3), (11), (13), 17, (18) & >wo (1), (3), (4), 6, 7, 8, 10, (11), 14, 15, (16), 17 & >LoVT 3, 6, (10), (11), 14, 15, 17 & >CLIP (3), (5), 6, 7, 8, 9, 10, (11), (13), 14, (15), 17 & best (11), 17
Uni-Xent $\tau' = 0.3$  & 4 (2) & 2 (2) & 9 (6) & 4 (3) & 7 (6) & 1 (1)  \\
% (2), (5), 9, 14 & 9, 14 & >/wo (1), (5), 6, 7, 8, 9, 14, (15), 17 & >LoVT (3), 6, 14, 17 & >CLIP 5, 7, 8, 9, 11, (13), 14 & best 14
Uni-Xent any $\tau'$    & -     & 7 (3) & 14 (8) & 7 (5) & 12 (9) & 3 (2) \\
% & (3), 9, (11), (13), 14, 17, (18) & >/wo (1), (3), (4), (5), 6, 7, 8, 9,  10, (11), 14, 15, (16), 17 & >LoVT 3, 6, (10), (11), 14, 15, 17 & > CLIP (3), 5, 6, 7, 8, 9, 10, 11, (13), 14, (15), 17 & best (11), 14, 17
\hline
Uni (any)               & -     & -     & 15 (12) & 9 (6) & 14 (10) & 9 (4) \\
% & & >/wo 1, (3), 4, 5, 6, 7, 8, 9, 10, (11), (12), 14, 15, 16, 17 & >LoVT 3, (5), 6, (10), (11), 14, 15, 16, 17 & > CLIP (1), (3), 5, 6, 7, 8, 9, 10, 11, (13), 14, (15), 16, 17 
\bottomrule
\end{tabular}

%% file: figures/results_rsna.tex
%auto-ignore
\begin{tabular}{lcccc|>{\columncolor{Gray}}c|cccc}
\toprule
& \multicolumn{3}{c}{RSNA YOLOv3 Finetune} & \multicolumn{3}{c}{RSNA YOLOv3 Frozen} & \multicolumn{3}{c}{RSNA Lin.\ Seg.} \\
         & \multicolumn{3}{c}{mAP ($\%$)} & \multicolumn{3}{c}{mAP ($\%$)} & \multicolumn{3}{c}{Dice ($\%$)} \\
                       & 1\% & 10\% & 100\% & 1\% & 10\% & 100\% & 1\% & 10\% & 100\% \\
    \midrule
%\multicolumn{10}{l}{\emph{General initialization methods}}\\
Random & 2.4$\pm$0.5 & 5.1$\pm$1.2 & 14.9$\pm$1.7 & 1.0$\pm$0.2 & 4.0$\pm$0.3 & 8.9$\pm$0.9 & 21.9$\pm$1.2 & 5.3$\pm$0.0 & 5.3$\pm$0.0 \\
ImageNet~\citep{ImageNet} & \textbf{5.0$\pm$0.7} & \textbf{12.4$\pm$0.8} & \textbf{19.0$\pm$0.2} & \textbf{3.6$\pm$1.4} & \textbf{8.0$\pm$0.1} & \textbf{15.7$\pm$0.3} & \textbf{27.5$\pm$0.6} & \textbf{38.3$\pm$0.0} & \textbf{43.3$\pm$0.0} \\
\midrule
%\multicolumn{10}{l}{\emph{Pre-Training on 30 \% of frontal MIMIC-CXR}}\\
CheXpert~\citep{chexpert} & \textbf{\underline{8.3$\pm$0.8}} & 12.4$\pm$1.6 & \underline{\textbf{21.3$\pm$0.3}} & 7.0$\pm$1.0 & 14.8$\pm$0.8 & 18.8$\pm$0.4 & 38.9$\pm$0.2 & 45.5$\pm$0.2 & 48.1$\pm$0.0 \\
%\_BYOL~\cite{BYOL} & 7.5$\pm$1.2 & 11.7$\pm$1.8 & 17.9$\pm$0.1 & 6.3$\pm$1.2 & 13.9$\pm$1.0 & \underline{\textbf{20.5$\pm$1.2}} & 43.3$\pm$0.1 & 48.9$\pm$0.1 & 50.9$\pm$0.0 \\
%\_SimCLR~\cite{SimCLR} & 6.4$\pm$0.7 & \underline{\textbf{14.9$\pm$0.6}} & 18.3$\pm$2.3 & 6.2$\pm$0.8 & 12.4$\pm$0.4 & 16.5$\pm$0.1 & 42.5$\pm$0.0 & 45.3$\pm$0.0 & 48.0$\pm$0.0 \\
BYOL~\citep{BYOL} & 7.0$\pm$1.0 & 11.9$\pm$1.1 & 18.8$\pm$0.2 & 9.6$\pm$0.2 & 14.0$\pm$1.2 & \textbf{21.0$\pm$0.2} & 42.9$\pm$0.1 & 47.8$\pm$0.2 & 50.0$\pm$0.0 \\
SimCLR~\citep{SimCLR} & 6.7$\pm$0.5 & \textbf{\underline{12.9$\pm$0.5}} & \dashuline{20.4$\pm$1.8} & 7.9$\pm$1.0 & 11.9$\pm$0.1 & 19.9$\pm$0.2 & 43.1$\pm$0.0 & 46.0$\pm$0.0 & 48.2$\pm$0.0 \\
PixelPro~\citep{PixelPro} & 4.8$\pm$0.6 & 12.6$\pm$1.2 & 19.8$\pm$0.4 & 3.1$\pm$0.2 & 6.4$\pm$0.5 & 13.4$\pm$0.3 & 25.9$\pm$0.2 & 34.6$\pm$0.0 & 39.8$\pm$0.1 \\
ConVIRT~\citep{ConVIRT} & 7.4$\pm$1.3 & 12.7$\pm$1.5 & 18.3$\pm$0.4 & \textbf{\underline{9.8$\pm$0.3}} & 14.8$\pm$1.1 & 8.4$\pm$1.1 & 42.1$\pm$0.1 & 47.1$\pm$0.2 & 50.2$\pm$0.0 \\
CLIP~\citep{CLIP}* & 7.2$\pm$0.8 & \dashuline{12.8$\pm$1.2} & 19.7$\pm$0.5 & 9.3$\pm$0.4 & 16.1$\pm$1.1 & 19.6$\pm$1.4 & 44.3$\pm$0.1 & 48.8$\pm$0.1 & 50.7$\pm$0.0 \\
%\rowcolor{LightCyan}
LoVT~\citep{LoVT} & \dashuline{7.7$\pm$1.0} & 11.7$\pm$0.5 & 17.2$\pm$1.3 & 8.6$\pm$1.5 & \textbf{\dashuline{17.9$\pm$0.4}} & 18.0$\pm$0.1 & \textbf{\underline{46.0$\pm$0.0}} & \textbf{\underline{49.4$\pm$0.0}} & \textbf{\underline{51.5$\pm$0.0}} \\
\midrule
LoVT /wo local & 5.3$\pm$0.9 & \textbf{12.5$\pm$1.3} & 19.9$\pm$0.0 & 6.8$\pm$0.8 & 17.4$\pm$0.9 & 18.1$\pm$0.1 & 44.6$\pm$0.1 & 48.7$\pm$0.0 & 51.2$\pm$0.0 \\
+ Uni-Gauss $\tau' = 0.2$ & \textbf{7.3$\pm$0.9} & 11.7$\pm$1.1 & 18.0$\pm$0.9 & 7.4$\pm$1.4 & \underline{\textbf{18.4$\pm$0.8}} & \textbf{\underline{22.1$\pm$0.1}} & 44.7$\pm$0.4 & 48.8$\pm$0.1 & 50.8$\pm$0.0 \\
+ Uni-Gauss $\tau' = 0.5$ & 6.4$\pm$0.8 & 10.6$\pm$0.9 & 19.5$\pm$0.2 & \textbf{8.3$\pm$1.3} & 17.6$\pm$0.1 & 19.1$\pm$1.0 & \textbf{\dashuline{45.9$\pm$0.0}} & \textbf{\underline{49.4$\pm$0.0}} & 50.8$\pm$0.0 \\
+ Uni-Xent $\tau' = 0.2$ & 5.8$\pm$1.2 & 10.8$\pm$1.4 & \textbf{20.0$\pm$1.0} & 7.0$\pm$1.1 & 17.4$\pm$1.4 & \dashuline{21.6$\pm$0.5} & 45.7$\pm$0.3 & 49.3$\pm$0.0 & 51.1$\pm$0.0 \\
+ Uni-Xent $\tau' = 0.3$ & 5.6$\pm$1.3 & 11.2$\pm$0.9 & 18.0$\pm$2.7 & 6.6$\pm$1.1 & 17.6$\pm$0.6 & 19.1$\pm$0.1 & 45.7$\pm$0.1 & 49.2$\pm$0.0 & \dashuline{\textbf{51.3$\pm$0.0}} \\
%\multicolumn{10}{l}{\emph{Pre-Training on 100 \% of frontal MIMIC-CXR}}\\
%CheXpert~\cite{chexpert} & \dashuline{10.0$\pm$1.9} & 12.4$\pm$0.9 & \textbf{\underline{22.2$\pm$0.4}} & 5.8$\pm$0.4 & 11.9$\pm$0.7 & 20.0$\pm$0.2 & 40.0$\pm$0.1 & 44.3$\pm$0.0 & 46.9$\pm$0.0 \\
%BYOL~\cite{BYOL} & 5.6$\pm$0.8 & 11.0$\pm$0.2 & 17.3$\pm$1.1 & 6.8$\pm$1.6 & 12.1$\pm$1.1 & 15.9$\pm$0.6 & 41.9$\pm$0.0 & 45.1$\pm$0.0 & 46.8$\pm$0.0 \\
%SimCLR~\cite{SimCLR} & 7.1$\pm$0.7 & 12.2$\pm$0.8 & 18.8$\pm$1.0 & 5.4$\pm$0.2 & 13.1$\pm$0.2 & 17.3$\pm$1.6 & 43.0$\pm$0.0 & 45.1$\pm$0.0 & 47.0$\pm$0.0 \\
%PixelPro~\cite{PixelPro} & 4.8$\pm$0.3 & 11.0$\pm$1.5 & 17.4$\pm$1.7 & 4.6$\pm$1.6 & 5.4$\pm$1.1 & 12.6$\pm$1.3 & 23.9$\pm$0.4 & 34.8$\pm$0.2 & 40.2$\pm$0.1 \\
%ConVIRT~\cite{ConVIRT} & \textbf{\underline{10.7$\pm$1.1}} & \textbf{\underline{13.3$\pm$0.8}} & 18.5$\pm$0.4 & 8.2$\pm$0.9 & 15.6$\pm$1.2 & 17.9$\pm$0.3 & 44.6$\pm$0.1 & 48.5$\pm$0.0 & 50.4$\pm$0.3 \\
%CLIP~\cite{CLIP}* & 7.0$\pm$1.5 & 10.7$\pm$1.1 & 19.9$\pm$0.8 & \textbf{\underline{11.9$\pm$0.7}} & 15.0$\pm$1.1 & 18.7$\pm$0.0 & 45.2$\pm$0.0 & 49.3$\pm$0.1 & 51.1$\pm$0.0 \\
%\rowcolor{LightCyan}
%LoVT (Ours) & 8.5$\pm$0.8 & \dashuline{13.2$\pm$0.6} & 18.1$\pm$3.2 & 9.6$\pm$1.2 & \textbf{\dashuline{16.4$\pm$1.3}} & \textbf{\dashuline{20.5$\pm$1.0}} & \textbf{\underline{46.3$\pm$0.0}} & \textbf{\underline{50.1$\pm$0.0}} & \textbf{\underline{51.8$\pm$0.0}} \\
\midrule
Task Nr. & 1 & 2 & 3 & 4 & 5 & 6 & 7 & 8 & 9 \\
\bottomrule
        \end{tabular}

%% file: figures/results_other.tex
%auto-ignore
\begin{tabular}{lccccccccc}
\toprule
& \multicolumn{3}{c}{COVID Rural} & \multicolumn{2}{c}{SIIM-ACR Pneumoth.} & \multicolumn{3}{c}{Object CXR} & NIH CXR \\
    & UNet & UNet & Linear & UNet & UNet & YOLOv3 & YOLOv3 & Linear & Linear \\
    & Finetune & Frozen & & Finetune & Frozen & Finetune & Frozen & & \\
    & Dice ($\%$) & Dice ($\%$) & Dice ($\%$) & Dice ($\%$) & Dice ($\%$) & fROC ($\%$) & fROC ($\%$) & Dice ($\%$) & \makecell{Avg \\ Dice ($\%$)} \\
    \midrule
%\multicolumn{10}{l}{\emph{General initialization methods}}\\
Random & 34.0$\pm$1.1 & 32.2$\pm$1.8 & 6.0$\pm$0.0 & 23.2$\pm$1.0 & 23.9$\pm$1.6 & 49.5$\pm$1.2 & 28.4$\pm$1.4 & 6.9$\pm$0.0 & 0.5$\pm$0.4 \\
ImageNet~\citep{ImageNet} & \textbf{43.9$\pm$2.0} & \textbf{41.9$\pm$1.7} & \textbf{32.6$\pm$0.7} & \textbf{38.5$\pm$0.9} & \textbf{36.9$\pm$0.7} & \textbf{62.5$\pm$0.4} & \textbf{52.7$\pm$1.3} & \textbf{37.8$\pm$0.0} & \textbf{2.6$\pm$1.6} \\
\midrule
%\multicolumn{10}{l}{\emph{Pre-Training on 30 \% of frontal MIMIC-CXR}}\\
CheXpert~\citep{chexpert} & 43.5$\pm$4.9 & 44.1$\pm$3.2 & 32.1$\pm$2.0 & 38.9$\pm$0.9 & 40.7$\pm$0.7 & 62.2$\pm$0.6 & 46.3$\pm$1.9 & 16.5$\pm$7.7 & 8.7$\pm$0.6 \\
BYOL~\citep{BYOL} & 46.2$\pm$1.6 & 47.5$\pm$1.6 & 36.9$\pm$1.7 & 43.1$\pm$0.6 & 42.9$\pm$0.3 & 59.6$\pm$1.0 & 55.7$\pm$1.0 & 32.3$\pm$0.1 & 6.0$\pm$0.1 \\
SimCLR~\citep{SimCLR} & 44.9$\pm$2.9 & 41.4$\pm$3.7 & 33.0$\pm$0.0 & 42.6$\pm$0.4 & 39.2$\pm$0.7 & 61.9$\pm$0.8 & 54.3$\pm$1.0 & 33.2$\pm$0.1 & 13.3$\pm$0.5 \\
%\_BYOL~\cite{BYOL} & 46.5$\pm$3.2 & 46.0$\pm$2.0 & 44.8$\pm$0.1 & 45.4$\pm$0.9 & 44.4$\pm$0.4 & 61.5$\pm$1.0 & 56.0$\pm$0.8 & 34.0$\pm$0.2 & 8.7$\pm$0.7 \\
%\_SimCLR~\cite{SimCLR} & 46.1$\pm$4.4 & 39.1$\pm$0.7 & 39.5$\pm$0.0 & 41.2$\pm$0.5 & 41.2$\pm$0.7 & 59.1$\pm$1.9 & 53.4$\pm$0.5 & 35.1$\pm$0.0 & 11.0$\pm$0.0 \\
PixelPro~\citep{PixelPro} & 47.0$\pm$3.4 & 38.5$\pm$3.9 & 26.6$\pm$0.4 & 39.3$\pm$0.8 & 39.1$\pm$0.3 & \textbf{63.1$\pm$0.7} & 46.3$\pm$0.2 & 29.9$\pm$0.2 & 1.8$\pm$0.0 \\
ConVIRT~\citep{ConVIRT} & 48.8$\pm$2.2 & 44.2$\pm$3.1 & 45.0$\pm$3.0 & 42.5$\pm$1.0 & 42.5$\pm$0.2 & 62.5$\pm$0.1 & 54.0$\pm$0.7 & 37.7$\pm$0.1 & 11.4$\pm$0.8 \\
CLIP~\citep{CLIP}* & 49.3$\pm$2.0 & 46.5$\pm$2.3 & \dashuline{46.2$\pm$0.3} & 42.8$\pm$1.5 & 42.5$\pm$0.6 & 62.9$\pm$0.8 & 55.5$\pm$2.1 & \textbf{39.0$\pm$0.0} & 12.5$\pm$1.0 \\
%\rowcolor{LightCyan}
LoVT~\citep{LoVT} & \textbf{49.5$\pm$1.3} & \textbf{\dashuline{49.2$\pm$4.6}} & \textbf{\underline{49.2$\pm$0.2}} & \textbf{\dashuline{43.4$\pm$0.7}} & \textbf{43.1$\pm$0.6} & 61.0$\pm$1.3 & \textbf{\dashuline{55.8$\pm$1.1}} & 37.6$\pm$0.2 & \textbf{\underline{13.4$\pm$0.8}} \\
\midrule
LoVT /wo local & 48.2$\pm$1.1 & 48.7$\pm$3.7 & 45.6$\pm$0.0 & \textbf{\underline{43.5$\pm$0.9}} & 42.6$\pm$0.4 & 60.4$\pm$1.6 & 54.8$\pm$1.1 & 38.5$\pm$0.2 & \textbf{\dashuline{13.2$\pm$0.9}} \\
+ Uni-Gauss $\tau' = 0.2$ & \textbf{\underline{51.4$\pm$3.1}} & 48.7$\pm$0.8 & \textbf{45.7$\pm$2.3} & 43.2$\pm$0.8 & 41.6$\pm$0.4 & 61.8$\pm$1.6 & 54.5$\pm$1.1 & \dashuline{39.1$\pm$0.0} & 11.7$\pm$1.3 \\
+ Uni-Gauss $\tau' = 0.5$ & 50.3$\pm$1.2 & 48.3$\pm$3.7 & 44.2$\pm$0.7 & 43.1$\pm$1.3 & \dashuline{44.3$\pm$0.4} & \textbf{\underline{63.4$\pm$0.8}} & \textbf{\underline{57.8$\pm$0.9}} & 38.9$\pm$0.0 & 8.9$\pm$1.6 \\
+ Uni-Xent $\tau' = 0.2$ & \dashuline{50.5$\pm$1.1} & \textbf{\underline{51.3$\pm$4.8}} & 44.0$\pm$0.0 & \dashuline{43.4$\pm$0.9} & 44.0$\pm$0.5 & \dashuline{63.2$\pm$1.3} & 54.9$\pm$1.9 & \textbf{\underline{39.8$\pm$0.0}} & 11.9$\pm$0.9 \\
+ Uni-Xent $\tau' = 0.3$ & 46.2$\pm$2.1 & 48.1$\pm$1.4 & 42.1$\pm$0.2 & 43.1$\pm$1.4 & \textbf{\underline{44.8$\pm$0.3}} & 60.7$\pm$1.2 & 52.2$\pm$2.4 & 38.7$\pm$0.1 & 8.4$\pm$1.2 \\
%\multicolumn{10}{l}{\emph{Pre-Training on 100 \% of frontal MIMIC-CXR}}\\
%CheXpert~\cite{chexpert} & 46.2$\pm$1.7 & 45.9$\pm$3.9 & 37.7$\pm$0.4 & 34.2$\pm$0.8 & 37.7$\pm$0.3 & 57.5$\pm$1.1 & 39.8$\pm$2.4 & 19.4$\pm$0.1 & \dashuline{15.2$\pm$0.0} \\
%BYOL~\cite{BYOL} & \dashuline{50.7$\pm$2.7} & 42.0$\pm$3.0 & 32.9$\pm$0.0 & 42.6$\pm$0.7 & 40.7$\pm$0.7 & 60.6$\pm$1.1 & 53.1$\pm$0.8 & 21.8$\pm$0.1 & 5.7$\pm$0.0 \\
%SimCLR~\cite{SimCLR} & 48.1$\pm$2.5 & 44.1$\pm$2.1 & 35.3$\pm$0.0 & 41.2$\pm$0.8 & 38.7$\pm$0.5 & 61.1$\pm$0.7 & 48.7$\pm$0.5 & 30.0$\pm$0.0 & 11.8$\pm$0.0 \\
%PixelPro~\cite{PixelPro} & 42.4$\pm$4.4 & 37.7$\pm$1.0 & 18.9$\pm$6.4 & 39.4$\pm$1.2 & 38.7$\pm$0.6 & \textbf{\underline{65.0$\pm$0.5}} & 46.2$\pm$1.2 & 29.7$\pm$0.1 & 1.8$\pm$0.0 \\
%ConVIRT~\cite{ConVIRT} & 47.9$\pm$0.7 & 46.0$\pm$1.1 & 42.7$\pm$2.0 & 39.3$\pm$0.3 & 43.1$\pm$0.3 & 60.6$\pm$1.2 & 52.5$\pm$1.0 & 36.0$\pm$0.0 & \textbf{\underline{18.6$\pm$0.1}} \\
%CLIP~\cite{CLIP}* & 48.6$\pm$2.4 & 45.8$\pm$4.1 & 41.7$\pm$0.1 & \dashuline{44.0$\pm$0.7} & \underline{\textbf{45.0$\pm$0.5}} & 62.8$\pm$0.5 & \dashuline{56.9$\pm$1.4} & \dashuline{39.4$\pm$0.0} & 11.4$\pm$0.8 \\
%\rowcolor{LightCyan}
%LoVT (Ours) & \textbf{\underline{51.2$\pm$2.5}} & \textbf{46.2$\pm$2.4} & \textbf{44.0$\pm$0.8} & \underline{\textbf{44.1$\pm$0.3}} & \dashuline{43.9$\pm$0.7} & 62.1$\pm$0.5 & \textbf{\underline{57.4$\pm$0.5}} & \textbf{\underline{39.9$\pm$0.0}} & 9.4$\pm$0.5 \\
\midrule
Task Nr. & 10 & 11 & 12 & 13 & 14 & 15 & 16 & 17 & 18 \\
\bottomrule
        \end{tabular}

%% file: ms.bbl
\begin{thebibliography}{19}
\providecommand{\natexlab}[1]{#1}
\providecommand{\url}[1]{\texttt{#1}}
\expandafter\ifx\csname urlstyle\endcsname\relax
  \providecommand{\doi}[1]{doi: #1}\else
  \providecommand{\doi}{doi: \begingroup \urlstyle{rm}\Url}\fi

\bibitem[Radford et~al.(2021)Radford, Kim, Hallacy, Ramesh, Goh, Agarwal,
  Sastry, Askell, Mishkin, Clark, Krueger, and Sutskever]{CLIP}
Alec Radford, Jong~Wook Kim, Chris Hallacy, Aditya Ramesh, Gabriel Goh,
  Sandhini Agarwal, Girish Sastry, Amanda Askell, Pamela Mishkin, Jack Clark,
  Gretchen Krueger, and Ilya Sutskever.
\newblock Learning transferable visual models from natural language
  supervision.
\newblock \emph{arXiv preprint arXiv: 2103.00020}, 2021.

\bibitem[Jia et~al.(2021)Jia, Yang, Xia, Chen, Parekh, Pham, Le, Sung, Li, and
  Duerig]{ALIGN}
Chao Jia, Yinfei Yang, Ye~Xia, Yi-Ting Chen, Zarana Parekh, Hieu Pham, Quoc Le,
  Yun-Hsuan Sung, Zhen Li, and Tom Duerig.
\newblock Scaling up visual and vision-language representation learning with
  noisy text supervision.
\newblock In \emph{ICML}, 2021.

\bibitem[Zhang et~al.(2020)Zhang, Jiang, Miura, Manning, and Langlotz]{ConVIRT}
Yuhao Zhang, Hang Jiang, Yasuhide Miura, Christopher~D. Manning, and Curtis~P.
  Langlotz.
\newblock Contrastive learning of medical visual representations from paired
  images and text.
\newblock \emph{arXiv preprint arXiv: 2010.00747}, 2020.

\bibitem[Müller et~al.(2022{\natexlab{a}})Müller, Kaissis, Zou, and
  Rueckert]{LoVT_MICCAI}
Philip Müller, Georgios Kaissis, Congyu Zou, and Daniel Rueckert.
\newblock Radiological reports improve pre-training for localized imaging tasks
  on chest x-rays.
\newblock In \emph{[to be published at] MICCAI}, 2022{\natexlab{a}}.

\bibitem[Müller et~al.(2022{\natexlab{b}})Müller, Kaissis, Zou, and
  Rueckert]{LoVT}
Philip Müller, Georgios Kaissis, Congyu Zou, and Daniel Rueckert.
\newblock Joint learning of localized representations from medical images and
  reports.
\newblock In \emph{[to be published at] ECCV, arXiv: 2112.02889},
  2022{\natexlab{b}}.

\bibitem[Liao et~al.(2021)Liao, Moyer, Cha, Quigley, Berkowitz, Horng, Golland,
  and Wells]{local_MI}
Ruizhi Liao, Daniel Moyer, Miriam Cha, Keegan Quigley, Seth Berkowitz, Steven
  Horng, Polina Golland, and William~M. Wells.
\newblock Multimodal representation learning via maximization of local mutual
  information.
\newblock In \emph{MICCAI}, 2021.

\bibitem[Wang and Isola(2020)]{uniformity}
Tongzhou Wang and Phillip Isola.
\newblock Understanding contrastive representation learning through alignment
  and uniformity on the hypersphere.
\newblock In \emph{ICML}, 2020.

\bibitem[Chen et~al.(2020)Chen, Kornblith, Norouzi, and Hinton]{SimCLR}
Ting Chen, Simon Kornblith, Mohammad Norouzi, and Geoffrey Hinton.
\newblock A simple framework for contrastive learning of visual
  representations.
\newblock In \emph{ICML}, 2020.

\bibitem[Johnson et~al.(2019{\natexlab{a}})Johnson, Pollard, Berkowitz,
  et~al.]{MIMIC-CXR}
A.~Johnson, T.~Pollard, S.~Berkowitz, et~al.
\newblock Mimic-cxr, a de-identified publicly available database of chest
  radiographs with free-text reports.
\newblock \emph{Sci Data}, 6\penalty0 (317), 2019{\natexlab{a}}.
\newblock \doi{https://doi.org/10.1038/s41597-019-0322-0}.

\bibitem[Johnson et~al.(2019{\natexlab{b}})Johnson, Pollard, Mark, Berkowitz,
  and Horng]{MIMIC-CXR-2}
A.~Johnson, T.~Pollard, R.~Mark, S.~Berkowitz, and S.~Horng.
\newblock Mimic-cxr database (version 2.0.0).
\newblock PhysioNet, 2019{\natexlab{b}}.

\bibitem[Johnson et~al.(2019{\natexlab{c}})Johnson, Lungren, Peng,
  et~al.]{MIMIC-CXR-JPG}
A.~Johnson, M.~Lungren, Y.~Peng, et~al.
\newblock Mimic-cxr-jpg - chest radiographs with structured labels (version
  2.0.0).
\newblock PhysioNet, 2019{\natexlab{c}}.

\bibitem[Desai and Johnson(2020)]{VirTex}
Karan Desai and Justin Johnson.
\newblock Virtex: Learning visual representations from textual annotations.
\newblock \emph{arXiv preprint arXiv: 2006.06666}, 2020.

\bibitem[Sariyildiz et~al.(2020)Sariyildiz, Perez, and Larlus]{ICMLM}
Mert Sariyildiz, Julien Perez, and Diane Larlus.
\newblock Learning visual representations with caption annotations.
\newblock In \emph{ECCV}, 2020.

\bibitem[He et~al.(2016)He, Zhang, Ren, and Sun]{ResNet50}
Kaiming He, Xiangyu Zhang, Shaoqing Ren, and Jian Sun.
\newblock Deep residual learning for image recognition.
\newblock In \emph{CVPR}, 2016.
\newblock \doi{10.1109/CVPR.2016.90}.

\bibitem[Devlin et~al.(2019)Devlin, Chang, Lee, and Toutanova]{BERT}
Jacob Devlin, Ming-Wei Chang, Kenton Lee, and Kristina Toutanova.
\newblock Bert: Pre-training of deep bidirectional transformers for language
  understanding.
\newblock In \emph{NAACL}, 2019.

\bibitem[Russakovsky et~al.(2015)Russakovsky, Deng, Su, Krause, Satheesh, Ma,
  Huang, Karpathy, Khosla, Bernstein, Berg, and Fei-Fei]{ImageNet}
Olga Russakovsky, Jia Deng, Hao Su, Jonathan Krause, Sanjeev Satheesh, Sean Ma,
  Zhiheng Huang, Andrej Karpathy, Aditya Khosla, Michael Bernstein,
  Alexander~C. Berg, and Li~Fei-Fei.
\newblock {ImageNet Large Scale Visual Recognition Challenge}.
\newblock \emph{IJCV}, 115\penalty0 (3):\penalty0 211--252, 2015.
\newblock \doi{10.1007/s11263-015-0816-y}.

\bibitem[Irvin et~al.(2019)Irvin, Rajpurkar, Ko, Yu, Ciurea-Ilcus, Chute,
  Marklund, Haghgoo, Ball, Shpanskaya, Seekins, Mong, Halabi, Sandberg, Jones,
  Larson, Langlotz, Patel, Lungren, and Ng]{chexpert}
Jeremy Irvin, Pranav Rajpurkar, Michael Ko, Yifan Yu, Silviana Ciurea-Ilcus,
  Chris Chute, Henrik Marklund, Behzad Haghgoo, Robyn Ball, Katie Shpanskaya,
  Jayne Seekins, {David A.} Mong, {Safwan S.} Halabi, {Jesse K.} Sandberg,
  Ricky Jones, {David B.} Larson, {Curtis P.} Langlotz, {Bhavik N.} Patel,
  {Matthew P.} Lungren, and {Andrew Y.} Ng.
\newblock Chexpert: A large chest radiograph dataset with uncertainty labels
  and expert comparison.
\newblock In \emph{AAAI}, pages 590--597, 2019.

\bibitem[Grill et~al.(2020)Grill, Strub, Altch\'{e}, Tallec, Richemond,
  Buchatskaya, Doersch, Avila~Pires, Guo, Gheshlaghi~Azar, Piot, kavukcuoglu,
  Munos, and Valko]{BYOL}
Jean-Bastien Grill, Florian Strub, Florent Altch\'{e}, Corentin Tallec, Pierre
  Richemond, Elena Buchatskaya, Carl Doersch, Bernardo Avila~Pires, Zhaohan
  Guo, Mohammad Gheshlaghi~Azar, Bilal Piot, koray kavukcuoglu, Remi Munos, and
  Michal Valko.
\newblock Bootstrap your own latent - a new approach to self-supervised
  learning.
\newblock In \emph{NeurIPS}, 2020.

\bibitem[Xie et~al.(2020)Xie, Lin, Zhang, Cao, Lin, and Hu]{PixelPro}
Zhenda Xie, Yutong Lin, Zheng Zhang, Yue Cao, Stephen Lin, and Han Hu.
\newblock Propagate yourself: Exploring pixel-level consistency for
  unsupervised visual representation learning.
\newblock \emph{arXiv preprint arXiv: 2011.10043}, 2020.

\end{thebibliography}
